\documentclass[10pt,twocolumn,letterpaper]{article}

\usepackage{graphicx}  
\usepackage{bm} 
\usepackage{bbding}
\usepackage{amsmath}

\usepackage{makecell}
\usepackage{booktabs}
\usepackage{multirow}
\usepackage{colortbl}
\usepackage{authblk}
\usepackage{pifont}

 \usepackage[pagenumbers]{cvpr} 

\definecolor{lightred}{RGB}{255,199,206}
\definecolor{lightblue}{RGB}{217, 231, 244}

\definecolor{cvprblue}{rgb}{0.21,0.49,0.74}
\usepackage[pagebackref,breaklinks,colorlinks,allcolors=cvprblue]{hyperref}

\def\paperID{} 
\def\confName{CVPR}
\def\confYear{2026}

\title{Customized Fusion: A Closed-Loop Dynamic Network for Adaptive Multi-Task-Aware Infrared-Visible Image Fusion}

\author{
	Zengyi Yang\textsuperscript{1}, 
	Yu Liu\textsuperscript{1}\thanks{Corresponding authors: Yu Liu and Huafeng Li.},
	Juan Cheng\textsuperscript{1},
	Zhiqin Zhu\textsuperscript{2},
	Yafei Zhang\textsuperscript{3},
	Huafeng Li\textsuperscript{3}\protect\footnotemark[1]
	\\
	\textsuperscript{1}Department of Biomedical Engineering, Hefei University of Technology,\\
	\textsuperscript{2}College of Automation, Chongqing University of Post and Telecommunications,\\
	\textsuperscript{3}Faculty of Information Engineering and Automation, Kunming University of Science and Technology \\
	\vspace{4pt}
	\tt\small zengyiyang0211@mail.hfut.edu.cn 
	\hspace{0.2cm} yuliu@hfut.edu.cn 
	\hspace{0.2cm} lhfchina99@kust.edu.cn
}

\begin{document}
\maketitle
\begin{abstract}
Infrared-visible image fusion aims to integrate complementary information for robust visual understanding, but existing fusion methods struggle with simultaneously adapting to multiple downstream tasks. To address this issue, we propose a Closed-Loop Dynamic Network (CLDyN) that can adaptively respond to the semantic requirements of diverse downstream tasks for task-customized image fusion. Specifically, CLDyN introduces a closed-loop optimization mechanism that establishes a semantic transmission chain to achieve explicit feedback from downstream tasks to the fusion network through a Requirement-driven Semantic Compensation (RSC) module. The RSC module leverages a Basis Vector Bank (BVB) and an Architecture-Adaptive Semantic Injection (A2SI) block to customize the network architecture according to task requirements, thereby enabling task-specific semantic compensation and allowing the fusion network to actively adapt to diverse tasks without retraining. To promote semantic compensation, a reward-penalty strategy is introduced to reward or penalize the RSC module based on task performance variations. Experiments on the M$^3$FD, FMB, and VT5000 datasets demonstrate that CLDyN not only maintains high fusion quality but also exhibits strong multi-task adaptability. The code is available at \url{https://github.com/YR0211/CLDyN}.
\end{abstract}    
\section{Introduction}
\label{sec:intro}

Infrared-visible image fusion aims to integrate complementary information from two modalities into a single image, thereby enhancing the overall scene representation \cite{21,22,23,24,25,26,27,29,30,54,55,58, 56, 57}. By combining the salient target cues from infrared images with the textural richness of visible images, the fused results exhibit improved semantic completeness and visual coherence, showing great potential in high-level vision tasks, including object detection, semantic segmentation, and salient object detection. This trend has motivated the development of task-aware image fusion methods that enable the fused results to align with the objectives of downstream tasks.

\begin{figure}[t!]
	\centering
	\includegraphics[width=0.48\textwidth]{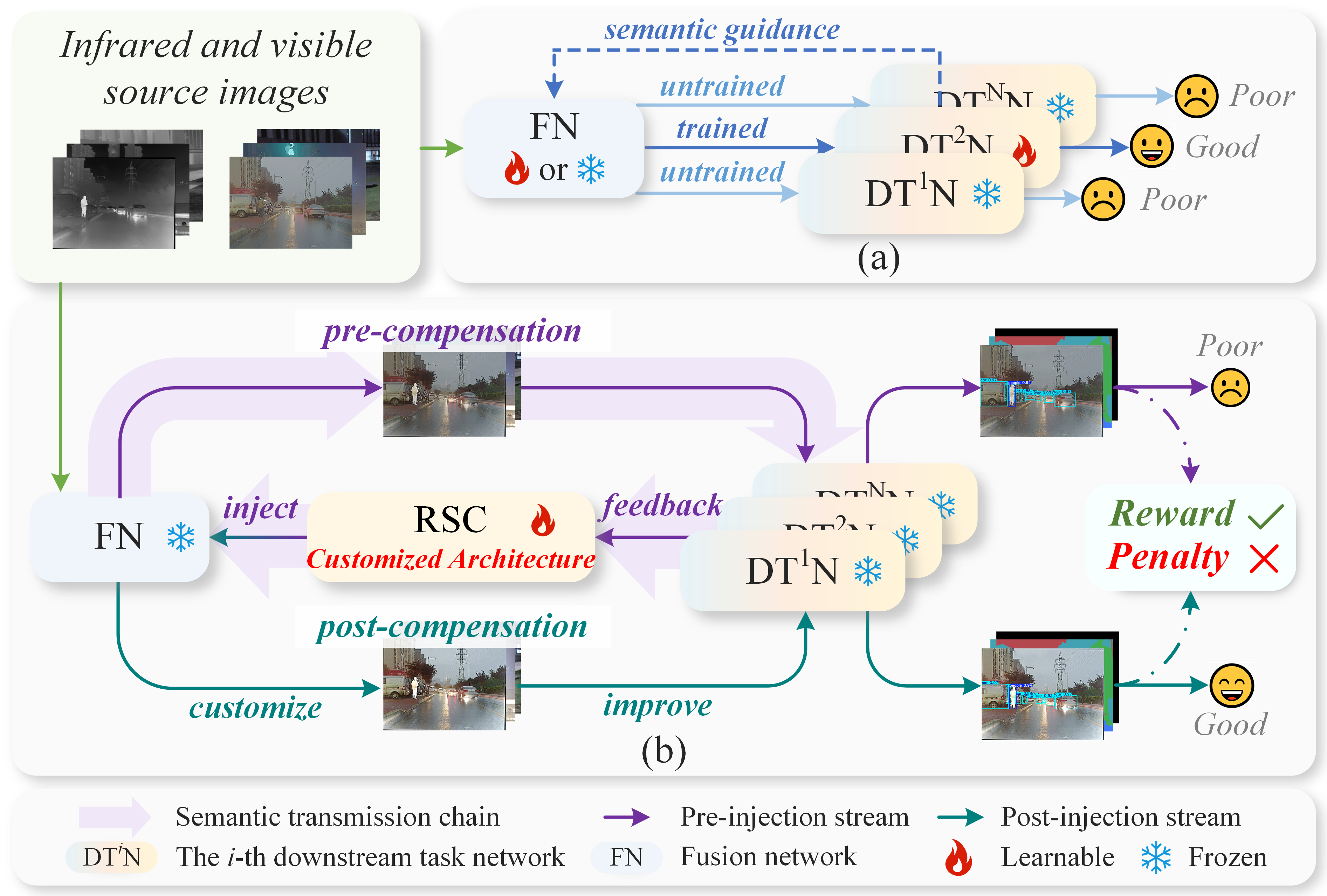}\vspace{-2mm}
	\caption{Comparison of processing paradigms between existing downstream task-aware image fusion methods (a) and the proposed closed-loop dynamic network (b).}
	\label{label1}\vspace{-5mm}
\end{figure}

Existing task-aware image fusion methods can be broadly grouped into two categories. The first category includes loss function-driven methods (e.g., SeAFusion \cite{1}, TDAL \cite{10}, MetaFusion \cite{2}, and TDFu \cite{4}), which design task-related loss functions to guide the fusion network toward learning semantically consistent representations. The second category includes task semantic-guided methods (e.g., DetFusion \cite{5}, UAAFusion \cite{6}, MRFS \cite{7}, and SAGE \cite{8}), which directly incorporate task features into the fusion process to enrich the semantic expressiveness of the fused images. Although these methods perform well on tasks whose downstream task networks (DTNs) are trained, their performance degrades on other tasks with untrained DTNs, as shown in Figure~\ref{label1}(a).

To overcome the above limitation, we propose a Closed-Loop Dynamic Network (CLDyN) for adaptive multi-task-aware infrared-visible image fusion via a closed-loop optimization mechanism. As illustrated in Figure~\ref{label1}(b), a semantic transmission chain is established among a fusion network (FN), a specially designed Requirement-driven Semantic Compensation (RSC) module, and multiple DTNs. As the core component of this chain, the RSC module receives semantic features fed back by different DTNs and performs task-specific semantic compensation on the FN accordingly. To achieve effective task-specific compensation, the RSC module integrates a Basis Vector Bank (BVB) and an Architecture-Adaptive Semantic Injection (A2SI) block. The BVB stores a set of learnable semantic bases optimized under multi-task constraints, supporting the generation of task-specific convolutional parameters. The A2SI block customizes the network architecture dynamically according to task requirements, facilitating more precise extraction of task-relevant semantic information. To achieve accurate semantic compensation, a reward-penalty strategy is introduced into the mechanism. By comparing the task performance before and after semantic compensation, this strategy rewards or penalizes the RSC module accordingly, allowing it to progressively acquire adaptive responsiveness to the semantic requirements of diverse tasks.

Overall, the proposed CLDyN introduces an adaptive task-customized paradigm for multi-task-aware image fusion. By incorporating a closed-loop optimization mechanism, the fusion network could adaptively adjust itself according to the semantic requirements of different DTNs without retraining, while the RSC module is trained only once to support multiple downstream tasks within a fixed task set. The main contributions of this work are threefold:

\begin{itemize}
	\item We introduce a closed-loop optimization mechanism for multi-task-aware infrared-visible image fusion. It establishes a semantic transmission chain between the fusion network and multiple downstream tasks, enabling dynamic semantic compensation according to task-specific requirements. Moreover, a reward-penalty strategy is introduced to achieve precise semantic compensation and effective multi-task adaptation.
	
	\item We propose an RSC module with task-customized architectures to perform task-specific semantic compensation through a Basis Vector Bank (BVB) and an Architecture-Adaptive Semantic Injection (A2SI) block, enabling semantic differentiation across multiple tasks.
	
	\item  Extensive experiments on the M$^3$FD, FMB, and VT5000 datasets demonstrate that our method achieves comprehensive adaptability across multiple downstream tasks, showcasing strong practical applicability.
\end{itemize}

\begin{figure*}[t!]
	\centering
	\includegraphics[width=0.97\textwidth]{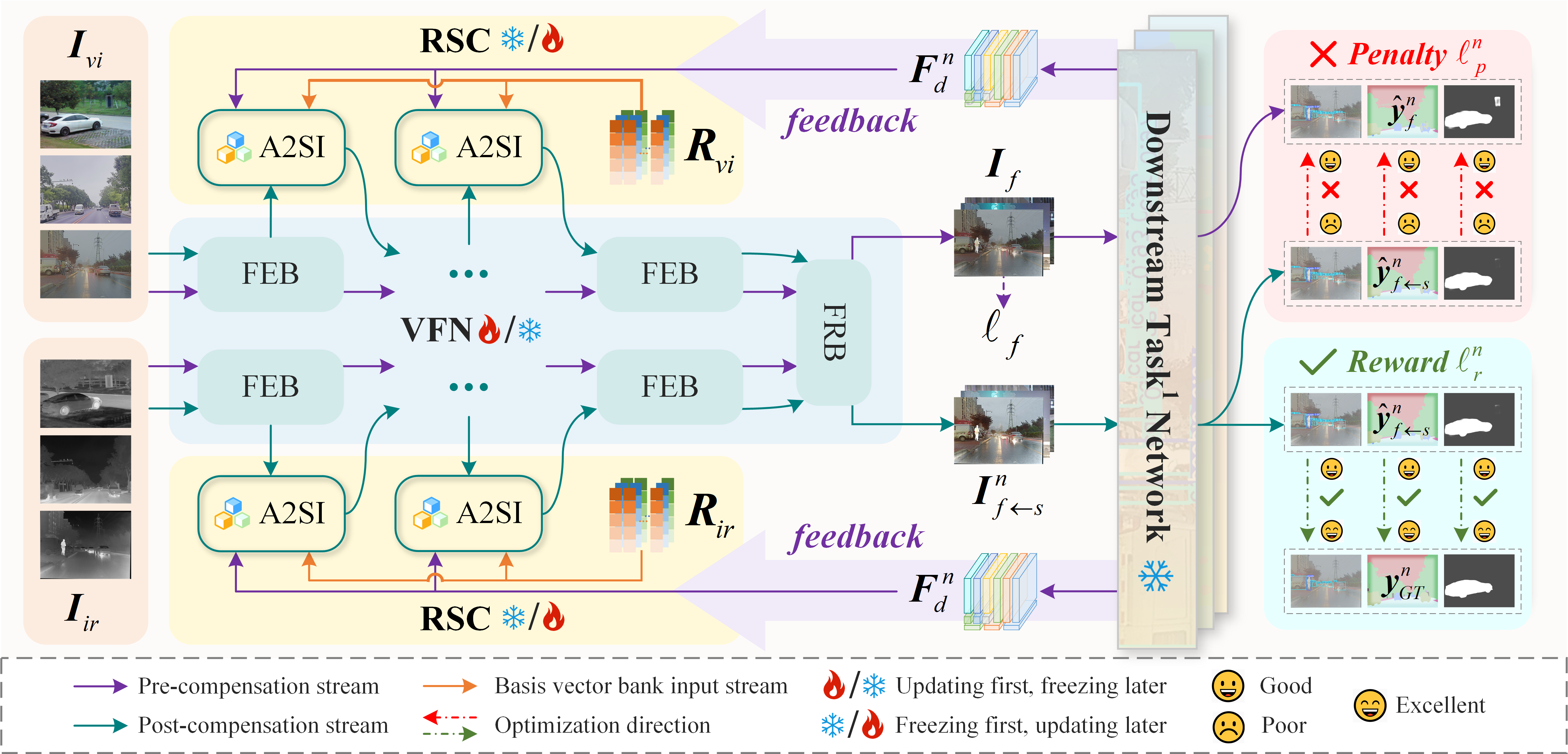}
	\caption{Overview of the adaptive multi-task-aware infrared-visible image fusion network. The network forms a semantic transmission chain, where semantic features from multiple downstream tasks guide the RSC module to perform task-specific compensation for the VFN. The reward-penalty strategy optimizes the compensation process by evaluating task performance before and after semantic compensation.}
	\label{label2}\vspace{-4mm}
\end{figure*}

\section{Related Work}
\label{sec:formatting}
In the study of task-aware image fusion, existing methods can generally be divided into two categories: loss function-driven methods and task semantic-guided methods.
\subsection{Loss Function-Driven Methods}
These methods constrain the fusion network with task-related losses to enhance semantic expressiveness, thus improving downstream task performance, as exemplified by SeAFusion \cite{1}, TDAL \cite{10}, and IRFS \cite{11}. However, the above methods often ignore feature-level semantic constraints. To address this limitation, MetaFusion \cite{2} aligns fusion and task features to enhance the task-relevant semantics of the fused images. To address the generalization degradation caused by noisy inputs, PAIF \cite{12} introduces adversarial perturbations and leverages task-related losses to learn more robust semantic representations.

In addition, several studies focus on balancing task performance and visual quality. BDLFusion \cite{13} alleviates task conflicts by randomly weighting the task-related and visual-related losses. To address optimization difficulties, TIMF \cite{14} combines architecture search with a parameter initialization strategy to accelerate semantic representation learning. Guided by task losses, TDFu \cite{4} learns weights to aggregate multi-source features for task-aware fusion. Although these methods improve task performance, most fail to exploit task features for semantic enhancement, thus limiting the semantic expressiveness of fused images.

\subsection{Task Semantic-Guided Methods}
These methods incorporate downstream task features to enhance the semantic of fused images. DetFusion \cite{5} and UAAFusion \cite{6} use task-related attention maps to guide fusion, while SMiF \cite{16} performs cross-attention between task and multi-source features for semantic injection. However, the large distribution gap between task and fusion features makes direct semantic injection suboptimal. To address this issue, methods such as PSFusion \cite{17}, MRFS \cite{7}, DetFusion++ \cite{18}, and SDCFusion \cite{19} enforce semantic consistency via feature sharing and joint optimization.

However, the injected semantics in these methods lack robustness across scenes. To address this issue, SAGE \cite{8} and SpTFuse \cite{20} inject semantic representations extracted by the Segment Anything Model (SAM) \cite{31} into the fused features to enhance generalization. These methods often perform well only on specific tasks and thus lack generality in multi-task scenarios. To overcome this limitation, IDF-TDDT \cite{9} fine-tunes the fusion network with task instructions for multi-task adaptation. Nevertheless, relying entirely on task instructions makes it difficult to capture task-specific semantics, leading to suboptimal performance. In addition, dynamic networks \cite{66, 67} and closed-loop control \cite{68, 69} have been widely studied for task adaptation.

In contrast, the CLDyN can explicitly leverage the semantic features fed back from multiple downstream tasks and adaptively perform task-specific semantic compensation on the fusion network, enabling multi-task adaptation.

\section{Methodology}
\subsection{Overview}
As illustrated in Figure \ref{label2}, the proposed method consists of two stages. In the first stage, a Vision-guided Fusion Network (VFN) is trained to produce visually pleasing fused images from multi-source inputs ${{\bm I}_{ir}}$ and ${{\bm I}_{vi}}$ using a series of Feature Extraction Blocks (FEB) and a Fusion Feature Reconstruction Block (FRB). 

In the second stage, the VFN is frozen, and a closed-loop optimization mechanism is introduced to enable dynamic multi-task adaptation. Through the semantic transmission chain, semantic features from multiple downstream tasks are propagated to the Requirement-driven Semantic Compensation (RSC) module, which performs task-specific compensation on the VFN to reconstruct fused images ${{\bm{I}}_{f \leftarrow s}^n}(n = 1, \ldots, N)$ adaptable to $N$ downstream tasks. The RSC module comprises a Basis Vector Bank (BVB) and an Architecture-Adaptive Semantic Injection (A2SI) block. The BVB provides learnable basis vectors to generate task-specific convolutional parameters, while the A2SI block adopts dynamic architectures to extract task-specific semantics effectively. To achieve precise semantic compensation, the  reward-penalty strategy evaluates task performance before and after compensation to constrain the RSC module.

\subsection{Vision-Guided Fusion Network}
As illustrated in Figure~\ref{label2}, the input infrared-visible images ${{\bm I}_{ir/vi}}$ are fed into $L$ FEBs for layer-by-layer feature extraction, producing multimodal features ${\bm F}_{ir/vi}^L$. These features are then passed into an FRB to reconstruct the fused image ${{\bm I}_f}$.  To ensure high visual quality, we introduce a fusion loss ${\ell _f}$ to constrain ${{\bm I}_f}$ at the pixel and gradient levels:
\begin{equation}\small
	{\ell _f} =
	{\left\| {{{\bm I}_f} - \max ({{\bm I}_{ir}}, {{\bm I}_{vi}})} \right\|_1}
	+\lambda {\left\| {\nabla {{\bm I}_f} - \max (\nabla {{\bm I}_{ir}}, \nabla {{\bm I}_{vi}})} \right\|_1},
\end{equation}
where $\lambda$ is a hyperparameter balancing the pixel and gradient losses. ${\|\cdot\|}_1$ denotes the $\ell_1$-norm and $\nabla$ denotes the Sobel gradient operator. This loss encourages the network to maintain pixel-level consistency while preserving texture details from the source images, therefore producing fused results with higher visual fidelity.

\subsection{Closed-Loop Optimization Mechanism}
Existing task-aware fusion methods often struggle to accurately respond to task-specific semantic requirements, limiting their capacity for multi-task adaptation. To address this limitation, we propose a Closed-Loop Optimization Mechanism, which consists of a semantic transmission chain and a reward-penalty strategy. Within this mechanism, the RSC module performs task-specific semantic compensation based on semantic features from different downstream tasks, allowing the VFN to adapt across diverse tasks.

\textbf{Semantic Transmission Chain.} As shown in Figure \ref{label2}, the multimodal images ${{\bm I}_{ir}}$ and ${{\bm I}_{vi}}$ are first fed into the frozen VFN to obtain multimodal features $\{{\bm F}_{ir/vi}^l \}_{l=1}^{L-1}$ and the initial fused result ${\bm I}_f$:
\begin{equation}\small
	({{\bm I}_f}, \{ {\bm F}_{ir/vi}^l \}_{l=1}^{L-1}) = {\rm VFN}({{\bm I}_{ir}}, {{\bm I}_{vi}}; \theta_\Phi^*),
	\label{eq:vfn_output}
\end{equation}
where $\theta_\Phi^*$ denotes the frozen parameters of the VFN, and ${\bm F}_{ir/vi}^l$ represents the output of the $l$-th FEB. ${\bm I}_f$ is then passed into the $n$-th task network ${\Gamma^n}$ to obtain the task output ${{\bm{\hat y}}_f^n}$ and the corresponding semantic features ${\bm F}_{d}^n$:
\begin{equation}\small
({{\bm{\hat y}}_f^n}, {\bm F}_{d}^n) = {\Gamma^n}({{\bm I}_f}; \theta_{\Gamma}^{n,*}), \quad n = 1, 2, \ldots, N,
\label{eq:task_output}
\end{equation}
where $\theta_{\Gamma}^{n,*}$ denotes the frozen parameters of ${\Gamma^n}$. The features ${\bm F}_{d}^n$ reflect the preference of the $n$-th task for specific structures, textures, or salient regions in the fused image.

After obtaining ${\bm F}_{d}^n$, the RSC module performs task-specific semantic compensation on ${\bm F}_{ir/vi}^l$ guided by ${\bm F}_{d}^n$, thus learning the consistency between task semantics and multimodal features to produce task-specific features as
\begin{equation}\small
\{ {\bm F}_{ir \leftarrow s/vi \leftarrow s}^{l,n} \}_{l=1}^{L-1} = {\rm RSC}(\{ {\bm F}_{ir/vi}^l \}_{l=1}^{L-1}, {{\bm F}_{d}^n}; \theta_\Psi),
\label{eq:rsc_compensation}
\end{equation}
where $\theta_\Psi$ denotes the trainable parameters of the RSC module. Through adaptive regulation, the RSC module emphasizes or suppresses features based on task requirements, for example, by highlighting thermal regions for detection tasks and enhancing edge structures for segmentation tasks, therefore aligning the fused features with task objectives at the semantic level. Finally, the semantically compensated features ${\bm F}_{ir \leftarrow s/vi \leftarrow s}^{l,n}$ are re-injected into the VFN to replace the original ${\bm F}_{ir/vi}^l$, thereby generating the task-customized fused images ${{\bm{I}}_{f \leftarrow s}^n}$ for the $n$-th task.
\begin{equation}\small
{{\bm{I}}_{f \leftarrow s}^n} = {\rm VFN}({{\bm I}_{ir}}, {{\bm I}_{vi}}; \theta_\Phi^*)
\Big|_{\{ {\bm F}_{ir/vi}^l \leftarrow {\bm F}_{ir \leftarrow s/vi \leftarrow s}^{l,n} \}}.
\label{eq:task_fusion}
\end{equation}
This process enables VFN to maintain semantic consistency across tasks, thus producing task-customized fusion results for multiple downstream tasks. During inference, the RSC module only receives semantic features from the downstream task network to perform adaptive semantic compensation for the VFN, enabling the fused results to meet the requirements of the corresponding tasks. Therefore, no gradient update is applied to any network during inference, and the RSC module introduces only 174.06G FLOPs. Moreover, the RSC module shares parameters across different downstream tasks, resulting in only 0.46M parameters.

\textbf{Reward-Penalty Strategy}. Although the semantic transmission chain can propagate features from different tasks, the RSC module may suffer from semantic compensation drift without effective constraints. To address this issue, we introduce a reward-penalty strategy that establishes an adaptive optimization process based on variations in task performance before and after semantic compensation.

Specifically, when the downstream task performance improves after semantic compensation, the RSC module is rewarded under Ground Truth (GT) supervision. Otherwise, an additional penalty is applied based on GT guidance to prevent invalid or harmful semantic drift. Accordingly, the reward loss $\ell_{r}^n$ and penalty loss $\ell_{p}^n$ are defined as follows:
\begin{equation}\small
	\begin{aligned}
		\ell_{r}^n &= c^n({{\bm{\hat{y}}}_{f \leftarrow s}^n}, {{\bm{y}}_{GT}^n}), \\
		\ell_{p}^n &= \max \big(0,\, c^n({{\bm{\hat{y}}}_{f \leftarrow s}^n}, {{\bm{y}}_{GT}^n}) 
		- c^n({{\bm{\hat y}}_f^n}, {{\bm{y}}_{GT}^n}) \big),
	\end{aligned}
	\label{eq:reward_loss}
\end{equation}
where $c^n(\cdot)$ denotes the loss function of the $n$-th downstream task, ${{\bm{\hat y}}_f^n}$ and ${{\bm{\hat{y}}}_{f \leftarrow s}^n}$ represent the task predictions before and after semantic compensation, respectively, 
and ${{\bm{y}}_{GT}^n}$ denotes the GT. The overall closed-loop optimization objective for the $n$-th downstream task is formulated as
\begin{equation}\small
	\ell_{cl}^n = \ell_{r}^n + \delta \ell_{p}^n,
	\label{eq:closed_loop_loss}
\end{equation}
where $\delta$ controls the penalty strength. Notably, we adopt CAGrad \cite{64} to mitigate gradient conflicts across tasks.

Through this reward-penalty strategy, the reward term guides perceptual alignment, while the penalty term suppresses excessive compensation. Consequently, the RSC module gradually develops a generalizable understanding of task-specific semantic requirements, thus enabling more efficient semantic compensation across multiple tasks.

\subsection{Requirement-Driven Semantic Compensation}
To enable the VFN to adapt to multiple downstream tasks, the RSC module performs task-specific semantic compensation on the features ${\bm F}_{ir/vi}^l$ extracted from the $l$-th FEB based on the semantic features ${{\bm F}_{d}^n}$ from the $n$-th downstream task. This process generates task-specific features ${\bm F}_{ir \leftarrow s/vi \leftarrow s}^{l,n}$, allowing the VFN to reconstruct fused images adaptable to diverse tasks. The RSC consists of a Basis Vector Bank (BVB) and $2(L-1)$ Architecture-Adaptive Semantic Injection (A2SI) blocks. The BVB stores learnable basis vectors to generate task-specific convolutional kernel parameters, while each A2SI block employs a task-customized dynamic network architecture to effectively extract the semantic information required by different tasks.

\begin{figure}[t!]
	\centering
	\includegraphics[width=0.48\textwidth]{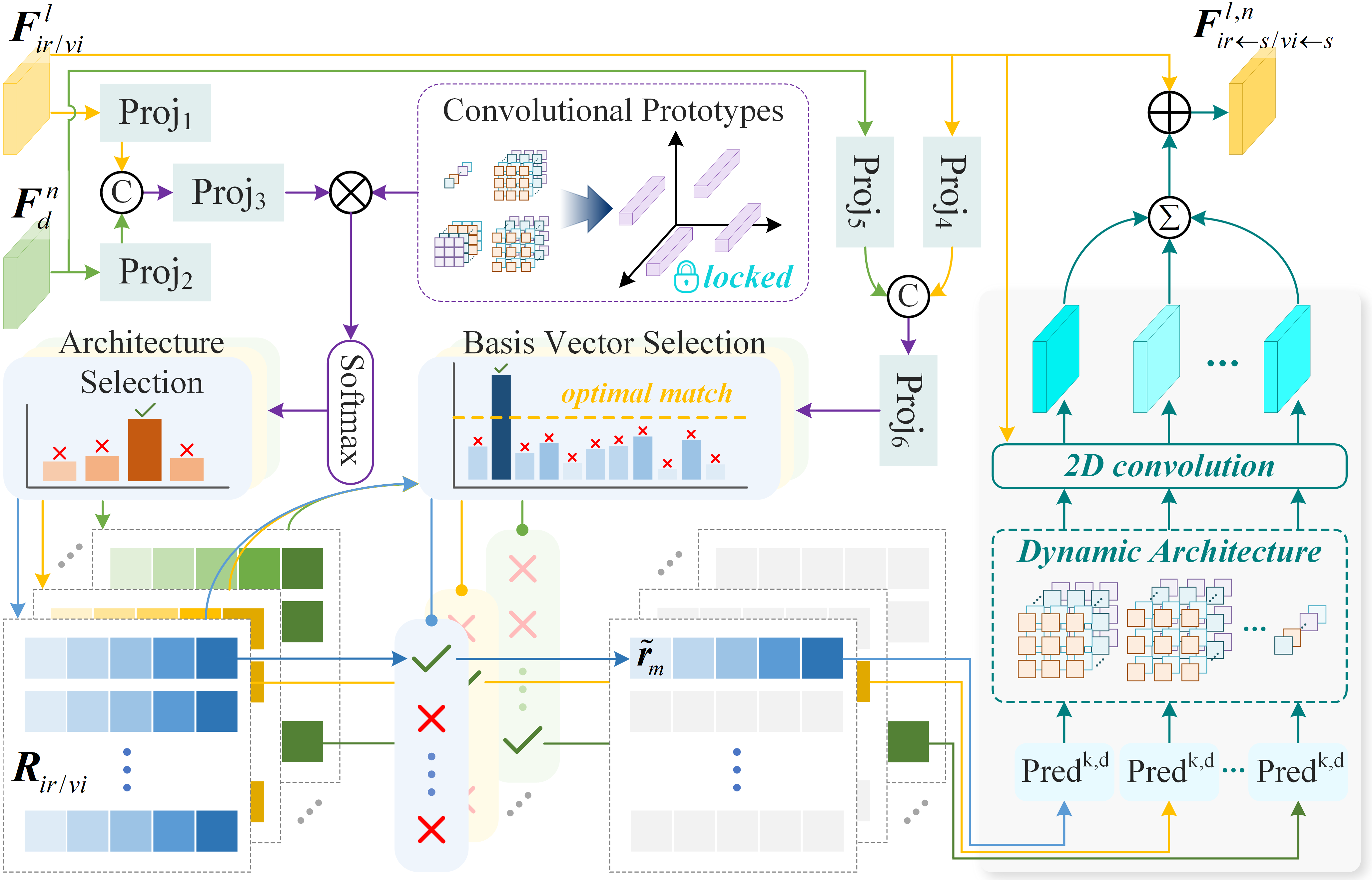}
	\caption{Architecture of the A2SI block. The A2SI block comprises six projection layers $Proj_i$ ($i = 1, \ldots, 6$) and several semantic extraction branches with dynamic convolutional architectures based on task semantics.}
	\label{label3}\vspace{-4mm}
\end{figure}

As illustrated in Figure~\ref{label3}, a single receptive field is insufficient to capture diverse task semantics. Therefore, $M$ architecture-adaptive semantic extraction branches are incorporated into each A2SI. Each branch adaptively adjusts its convolutional configuration guided by ${\bm F}_{ir/vi}^l$ and ${{\bm F}_{d}^n}$, thus capturing task-specific semantic information at multiple receptive fields. To realize structural adaptivity, we introduce orthogonal convolutional prototypes ${\bm p} \in \mathbb{R}^{4 \times e_1}$, which represent four typical convolutional configurations as ${\bm p} = [{{\bm p}_{1,1}}, {{\bm p}_{3,1}}, {{\bm p}_{3,2}}, {{\bm p}_{3,3}}]$. 
Here, ${{\bm p}_{k,d}}$ denotes a prototype with kernel size $k \times k$ and dilation rate $d$, and $e_1$ denotes the embedding dimension, where $e_1 = 16$. All prototypes are mutually orthogonal and remain frozen to preserve independence among configuration representations.

To adaptively determine the convolutional configuration for each branch, ${\bm F}_{ir/vi}^l$ and ${{\bm F}_{d}^n}$ are first fed into two projection layers $Proj_{1/2}$ for feature projection. Specifically, $Proj_{1/2}$ consists of two convolutional layers. Their outputs are concatenated and then processed by $Proj_3$ (two linear layers) for information aggregation. The aggregated feature is multiplied by the prototypes ${\bm p}$ and normalized by the Softmax function to produce the convolution configuration selection matrix for $M$ branches:
\begin{equation}\small
	{\bm S} = {\rm Softmax}\big({\bm p} \times {\rm Resh}(Proj_3([Proj_1({\bm F}_{ir/vi}^l), Proj_2({{\bm F}_{d}^n})]))\big),
	\label{eq:s_config}
\end{equation}
where ${\bm S}(j,i)$ denotes the probability that the $i$-th branch selects the $j$-th convolutional configuration. 
The configuration with the highest probability is chosen for that branch. 
The operation ${\rm Resh}(\cdot): \mathbb{R}^{1 \times e_1 M} \to \mathbb{R}^{e_1 \times M}$ 
represents tensor reshaping, and $[ \cdot ]$ indicates channel concatenation. Through this design, each A2SI module adaptively customizes the optimal convolutional structures based on task-specific semantics and the scene context, thereby enhancing adaptability to diverse task requirements.

After determining the convolutional configurations of multiple branches, we introduce the BVB ${{\bm R}_{ir/vi}} = [ {\bm R}_{ir/vi}^{1,1}, {\bm R}_{ir/vi}^{3,1}, {\bm R}_{ir/vi}^{3,2}, {\bm R}_{ir/vi}^{3,3} ] \in \mathbb{R}^{4 \times 32 \times e_2}$ to predict kernel parameters. The BVB contains four sub-banks for the convolutional configurations. Each sub-bank ${\bm R}_{ir/vi}^{k,d} = [{\bm r}_{ir/vi,1}^{k,d}, \ldots, {\bm r}_{ir/vi,32}^{k,d}] \in \mathbb{R}^{32 \times e_2}$ includes 32 learnable basis vectors of dimension $e_2$, initialized as pairwise orthogonal vectors, where $e_2 = 256$. Trained with the reward-penalty strategy, ${\bm r}_{ir/vi,i}^{k,d}$ learns fundamental semantic representations that provide diverse priors for kernel generation.

To select the most suitable basis vector according to task semantics, ${\bm F}_{ir/vi}^l$ and ${{\bm F}_{d}^n}$ are fed into $Proj_{4}$ and $Proj_{5}$ to extract semantic features, then concatenated and are aggregated via $Proj_{6}$. We then compute the cosine similarity between the aggregated features and each basis vector:
\begin{equation}\small
	{s_i} = \cos(Proj_6([Proj_4({\bm F}_{ir/vi}^l), Proj_5({{\bm F}_{d}^n})]), {\bm r}_{ir/vi,i}^{k,d}),
	\label{eq:cos_similarity}
\end{equation}
where ${s_i}$ denotes the similarity score, and $\cos(\cdot,\cdot)$ is the cosine similarity function. $Proj_{4/5/6}$ share the same architecture as $Proj_{1/2/3}$. According to the maximum similarity principle, the most relevant basis vector is selected as
\begin{equation}\small
	{{\tilde{\bm r}}_m} = {\bm r}_{ir/vi,i^*}^{k,d}, 
	\quad i^* = \mathop{\arg\max}\limits_{i \in \{1, \ldots, 32\}} (s_i).
	\label{eq:basis_selection}
\end{equation}
This process leverages both task semantics and scene context to select ${{\tilde{\bm r}}_m}$, enabling the A2SI block to generate robust convolutional parameters across different tasks.

Once the optimal basis vector ${{\tilde{\bm r}}_m}$ is obtained, it is fed into the configuration-specific prediction block Pred$^{k,d}$ to generate the convolutional parameters for the $m$-th branch:
\begin{equation}\small
	{\bm W}_m^{k,d} = {\rm Pred}^{k,d}({{\tilde{\bm r}}_m}), 
	\quad m = 1, 2, \ldots, M,
	\label{eq:kernel_generation}
\end{equation}
where Pred$^{k,d}$ consists of two linear layers. The predicted kernels ${\bm W}_m^{k,d}$ are then applied in parallel convolutional operations to extract task-specific semantic information across multiple receptive fields. The outputs of all branches are aggregated and injected back into the original features, producing task-specific features:
\begin{equation}\small
	{\bm F}_{ir \leftarrow s/vi \leftarrow s}^{l,n} = {\bm F}_{ir/vi}^l + \frac{1}{M}\sum\limits_{m=1}^M ({\bm W}_m^{k,d} * {\bm F}_{ir/vi}^l),
	\label{eq:semantic_injection}
\end{equation}
where $*$ denotes the convolution operation.

During this process, the A2SI block generates task-specific features ${\bm F}_{ir \leftarrow s/vi \leftarrow s}^{l,n}$, which replace the original ${\bm F}_{ir/vi}^l$ and are propagated through the subsequent FEBs, progressively transmitting task-specific semantics across layers. Finally, the VFN reconstructs the fused image ${{\bm{I}}_{f \leftarrow s}^n}$ adapted to the $n$-th task. Under the closed-loop optimization mechanism, the RSC module completes an adaptive task-specific semantic compensation process, thereby improving the adaptability of the VFN across multiple tasks.
\section{Experiments}
\subsection{Implementation Details}

\begin{table*}[htbp]
	\centering
	\caption{Quantitative comparison between the proposed method and existing state-of-the-art approaches. The best and second-best performances for each metric are highlighted with {\setlength{\fboxsep}{2.3pt}\colorbox{lightred}{Red}} and {\setlength{\fboxsep}{2.3pt}\colorbox{lightblue}{Blue}} backgrounds, respectively.}\vspace{-2mm}
	\renewcommand\arraystretch{1.4}
	{\footnotesize\centerline{\tabcolsep=2.9pt
			\begin{tabular}{c|ccccc|ccccc|ccccc}
				\toprule
				\textbf{Datasets} & \multicolumn{5}{c|}{\textbf{M$^3$FD}}             & \multicolumn{5}{c|}{\textbf{FMB}}              & \multicolumn{5}{c}{\textbf{VT5000}} \\
				\midrule
				\textbf{Methods} & $\rm{MI} \uparrow$ & ${Q_{AB/F}} \uparrow$  & ${Q_{CB}} \uparrow$   & ${Q_{CV}} \downarrow$   & ${Q_{CC}} \uparrow$    & $\rm{MI}$ & ${Q_{AB/F}}$  & ${Q_{CB}}$   & ${Q_{CV}}$   & ${Q_{CC}}$    & $\rm{MI}$ & ${Q_{AB/F}}$  & ${Q_{CB}}$   & ${Q_{CV}}$   & ${Q_{CC}}$ \\
				\midrule
				CoCo\cite{51} & 1.9284  & 0.3980  & 0.4004  & 813.92  & 0.6101  & 2.1332  & 0.3857  & 0.3893  & 615.32  & \cellcolor{lightred}{0.6834} & 1.9208  & 0.3660  & 0.4270  & 788.37  & \cellcolor{lightblue}{0.7075} \\
				TDAL\cite{10} & 1.4305  & 0.3047  & 0.3605  & 1073.90  & 0.5077  & 1.4922  & 0.3306  & 0.3367  & 740.31  & 0.5818  & 1.6225  & 0.3071  & 0.4193  & 981.00  & 0.6396  \\
				IRFS\cite{11}  & 1.9552  & 0.5165  & 0.3996  & 719.71  & 0.5578  & 2.1391  & 0.5470  & 0.3874  & 462.69  & 0.6248  & 1.9670  & 0.4735  & 0.4773  & 523.45  & 0.6980  \\
				SMiF\cite{16} & 2.1313  & \cellcolor{lightblue}{0.6601} & \cellcolor{lightblue}{0.4683} & \cellcolor{lightblue}{488.67} & 0.6192  & 2.2615  & \cellcolor{lightblue}{0.6924} & \cellcolor{lightblue}{0.4650} & \cellcolor{lightblue}{326.10} & 0.6743  & 1.7521  & 0.5059  & 0.4925  & 472.46  & 0.6823  \\
				MRFS\cite{7}  & 2.2639  & 0.5540  & 0.4218  & 633.00  & \cellcolor{lightblue}{0.6370} & 2.3991  & 0.6230  & 0.4329  & 433.04  & 0.6701  & 1.9274  & 0.5150  & 0.4507  & 416.98  & 0.6957  \\
				TIMF\cite{14} & \cellcolor{lightred}{2.5942} & 0.4938  & 0.3405  & 880.86  & 0.6325  & \cellcolor{lightblue}{2.4035} & 0.5023  & 0.3180  & 748.51  & 0.6532  & \cellcolor{lightred}{2.5749} & 0.5196  & \cellcolor{lightblue}{0.5139} & 591.51  & 0.6947  \\
				SAGE\cite{8}  & 2.1834  & 0.5903  & 0.4386  & 508.77  & 0.6173  & 2.3767  & 0.6372  & 0.4421  & 333.38  & 0.6767  & 2.0776  & \cellcolor{lightblue}{0.5249} & 0.4604  & \cellcolor{lightblue}{392.64} & 0.7054  \\
				
				Ours  & \cellcolor{lightblue}{2.5265} & \cellcolor{lightred}{0.6900} & \cellcolor{lightred}{0.4729} & \cellcolor{lightred}{472.62} & \cellcolor{lightred}{0.6385} & \cellcolor{lightred}{2.6219} & \cellcolor{lightred}{0.7124} & \cellcolor{lightred}{0.4736} & \cellcolor{lightred}{316.41} & \cellcolor{lightblue}{0.6831} & \cellcolor{lightblue}{2.3352} & \cellcolor{lightred}{0.6519} & \cellcolor{lightred}{0.5414} & \cellcolor{lightred}{331.15} & \cellcolor{lightred}{0.7091} \\
				\bottomrule
	\end{tabular}}}\vspace{-4mm}
	\label{tab:1}%
\end{table*}%

The overall training process consists of two stages. In the first stage, we randomly select infrared-visible image pairs from LLVIP \cite{33} (201 pairs), MSRS \cite{34} (217 pairs), RoadScene \cite{35,36} (200 pairs), and M$^3$FD (the part for fusion) \cite{10} (230 pairs) for train the VFN. To increase data diversity, random rotation, horizontal flipping, and random cropping into $256\times256$ patches are applied. In the second stage, the VFN is frozen, and only the RSC module is trained and evaluated using image pairs from M$^3$FD (the part for detection) (train: 3150, test: 525), FMB \cite{16} (train: 1220, test: 280), and VT5000 \cite{32} (train: 2500, test: 2500). Data augmentation strategies follow those in YOLOv5, SegFormer \cite{37}, and CTDNet \cite{38} to further enrich the training data. Notably, we actually adopt the official fixed data split if it is provided by a dataset (e.g., FMB and VT5000) and follow the common practice \cite{59, 60} if not (e.g., M$^3$FD).

During training, both stages adopt the Adam \cite{39} optimizer with batch sizes of 16 and 4, and initial learning rates of $1\times10^{-3}$ and $1\times10^{-2}$, respectively. The total training epochs are 100 and 50 for the two stages. The main hyperparameters $L$, $\delta$, and $M$ are set to 2, 5, and 4, respectively. For training and evaluation, we employ YOLOv5s, SegFormer (mit-b2), and CTDNet-18 as downstream task networks. All experiments are implemented in PyTorch and conducted on a single NVIDIA GeForce RTX 4090 GPU.

\subsection{Fusion Performance Comparison}
To evaluate the fusion performance of CLDyN, we compare it with seven representative fusion methods, including CoCo \cite{51}, TDAL \cite{10}, IRFS \cite{11}, SMiF \cite{16}, MRFS \cite{7}, TIMF \cite{14}, and SAGE \cite{8}. Notably, the fusion networks of these methods are not retrained, and the officially released models are used. A quantitative analysis is conducted using five widely adopted metrics in image fusion \cite{53}: Mutual Information ($\rm{MI}$) \cite{40,45}, Gradient-based Fusion Performance (${Q_{AB/F}}$) \cite{41}, Chen-Blum (${Q_{CB}}$) \cite{43}, Chen-Varshney (${Q_{CV}}$) \cite{44}, and Correlation Coefficient (${Q_{CC}}$) \cite{42}. As shown in Table \ref{tab:1}, the proposed method consistently ranks first across most evaluation metrics, demonstrating its significant advantage in fusion performance. It is worth noting that the visually dominant fused images before semantic compensation are used for comparison.

\subsection{Multi-Task Performance Comparison}
Existing fusion methods designed to enhance downstream task performance can generally be categorized into two types: task network retraining and joint training methods. Therefore, we compare the proposed method with the above methods to verify its superiority in multi-task adaptability across \textbf{Object Detection (OD)}, \textbf{Semantic Segmentation (Seg)}, and \textbf{Salient Object Detection (SOD)} tasks.

\begin{figure*}[t!]
	\centering
	\includegraphics[width=0.98\textwidth]{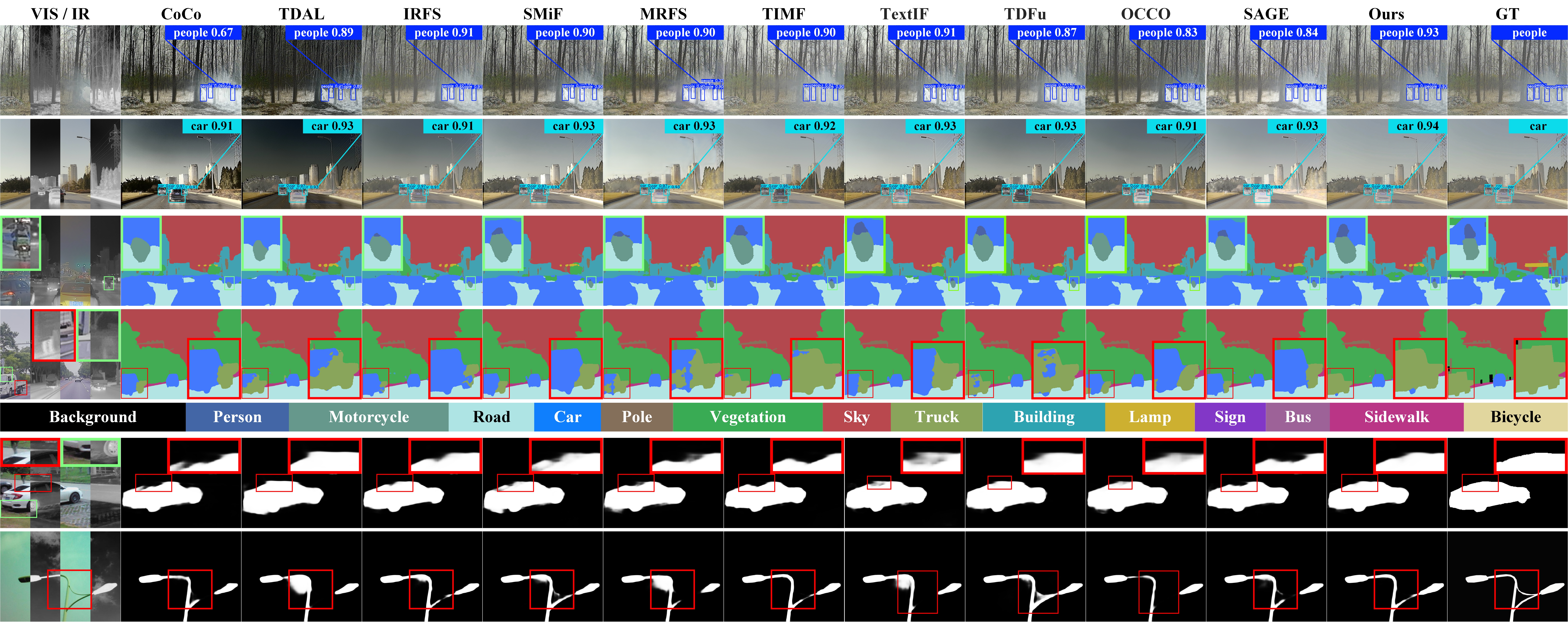}\vspace{-2mm}
	\caption{Qualitative comparison between the proposed method and the “task network retraining” methods.}\vspace{-2mm}
	\label{label5}
\end{figure*}

\textbf{Comparison with “Task Network Retraining” Methods}. We compare our method with “task network retraining” methods, which retrain downstream task networks using the fused results to improve task performance. Specifically, we retrain the task network on the fused outputs of ten representative fusion methods (CoCo \cite{51}, TDAL \cite{10}, IRFS \cite{11}, SMiF \cite{16}, MRFS \cite{7}, TIMF \cite{14}, TextIF \cite{61}, TDFusion \cite{4}, OCCO \cite{62}, and SAGE \cite{8}) for comparison. As shown in Table \ref{tab:2}, the proposed method achieves the highest scores on most evaluation metrics while requiring the fewest trainable parameters and the lowest FLOPs in the trainable components. Even when ranking second, its results remain very close to the best results (e.g., mIoU and E$_m$). In contrast, the compared methods fail to achieve comprehensive multi-task adaptability and require higher computational and parameter costs. As shown in Figure \ref{label5}, qualitative results further demonstrate that the results of our method are closer to the GT across multiple tasks. Notably, the FLOPs are calculated for the trainable components using input images with a resolution of $600 \times 800$.

\begin{table}[htbp]
	\centering
	\vspace{-2.5mm}
	\caption{Quantitative comparison of the proposed method with the “task network retraining” methods. The number of parameters and the computational cost of the trainable parts are reported in the table. The best and second-best performances for each metric are highlighted with {\setlength{\fboxsep}{2.3pt}\colorbox{lightred}{Red}} and {\setlength{\fboxsep}{2.3pt}\colorbox{lightblue}{Blue}} backgrounds.}\vspace{-2mm}
		\renewcommand\arraystretch{1.4}
	{\footnotesize\centerline{\tabcolsep=2.5pt
	\begin{tabular}{c|c|c|cc|c|c}
		\toprule
		\multirow{2}{*}{\textbf{Methods}} & \textbf{OD}    & \textbf{Seg}   & \multicolumn{2}{c|}{\textbf{SOD}} & \multicolumn{1}{c|}{\multirow{2}{*}{\thead{\textbf{Params}\\(M)}}} & \multirow{2}{*}{\thead{\textbf{FLOPs}\\(G)}} \\
		\cline{2-5}          & mAP$_{50 \rightarrow 95} \uparrow$   & mIoU $\uparrow$  & mF$_{\beta} \uparrow$    & E$_m \uparrow$    &       &  \\
		\midrule
		CoCo\cite{51} & 0.6080  & 58.95  & 0.8020  & 0.9020  &  \cellcolor{lightblue}{}  &  \cellcolor{lightblue}{} \\
		TDAL\cite{10} & 0.6157  & 59.80  & 0.7977  & 0.8993  &    \cellcolor{lightblue}{}   & \cellcolor{lightblue}{} \\
		IRFS\cite{11}  & \cellcolor{lightblue}{0.6306} & 59.43  & \cellcolor{lightblue}{0.8114} & \cellcolor{lightred}{0.9091} &   \cellcolor{lightblue}{}   & \cellcolor{lightblue}{} \\
		SMiF\cite{16} & 0.6273  & 59.15  & 0.8023  & 0.9037  &   \cellcolor{lightblue}{}    & \cellcolor{lightblue}{} \\
		MRFS\cite{7}  & 0.6189  & 58.28  & 0.8032  & 0.9026  &    \cellcolor{lightblue}{}    & \cellcolor{lightblue}{}\\
		TIMF\cite{14} & 0.6166  & \cellcolor{lightred}{60.86} & 0.7985  & 0.8998  &  \cellcolor{lightblue}{46.52}    & \cellcolor{lightblue}{183.82} \\
		
		TextIF \cite{61} & 0.6219  & 59.36 & 0.8033 & 0.9028 &  \cellcolor{lightblue}{}     & \cellcolor{lightblue}{}\\
		TDFu \cite{4} & 0.6287  & 60.12 & 0.7998  & 0.9003 &  \cellcolor{lightblue}{}     & \cellcolor{lightblue}{} \\
		OCCO \cite{62}  & \cellcolor{lightred}{0.6320} & 58.57 & 0.8030  & 0.9017 &  \cellcolor{lightblue}{}     & \cellcolor{lightblue}{} \\
		
		SAGE\cite{8}  & 0.6225  & 54.89  & 0.8093  & 0.9066  &   \cellcolor{lightblue}{}    & \cellcolor{lightblue}{} \\
		
		Ours  & 0.6304 & \cellcolor{lightblue}{60.34} & \cellcolor{lightred}{0.8129} & \cellcolor{lightblue}{0.9087} & \cellcolor{lightred}{0.46} & \cellcolor{lightred}{174.06} \\
		\bottomrule
	\end{tabular}}}\vspace{-3mm}
	\label{tab:2}%
\end{table}%

\begin{figure}[t!]
	\centering
	\includegraphics[width=0.47\textwidth]{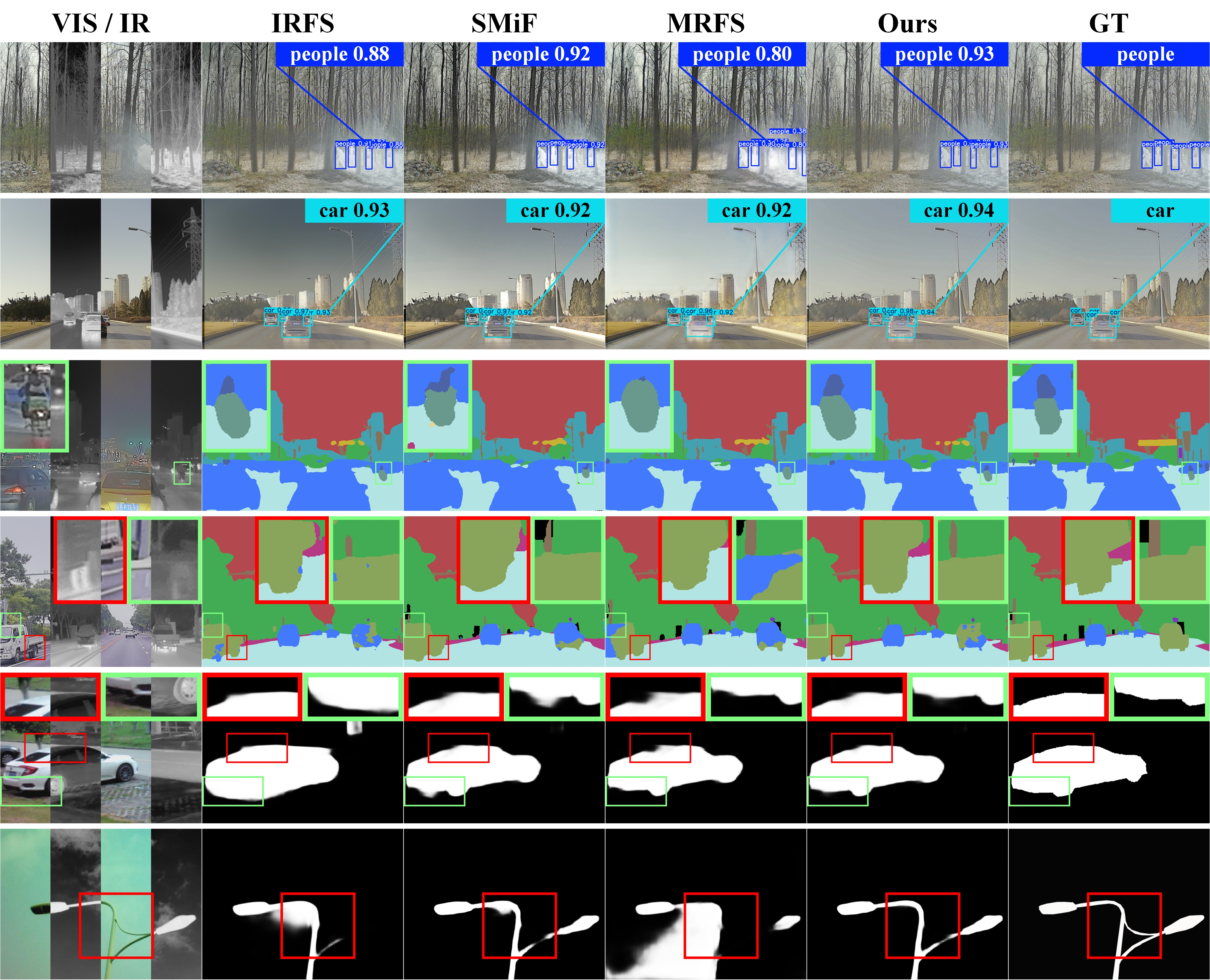}\vspace{-2mm}
	\caption{Qualitative comparison between the proposed method and the “joint training” methods.}\vspace{-4mm}
	\label{label6}
\end{figure}

\textbf{Comparison with “Joint Training” Methods}. We further compare our method with "joint training" methods, which cascade the fusion network with task networks and jointly optimize them with fusion and task losses. Representative methods include IRFS, SMiF, and MRFS, where IRFS targets SOD, while SMiF and MRFS target Seg. As shown in Table \ref{tab:3}, the compared methods achieve competitive performance only on the tasks participating in the training, but degrade significantly on others. In contrast, our method maintains consistently superior performance across multiple tasks, with only marginal gaps compared to the compared methods on their trained tasks. As shown in Figure \ref{label6}, qualitative results further support these findings. Moreover, our method requires substantially fewer parameters and a lower FLOPs for multi-task adaptation than "joint training" methods do for single-task adaptation.

\begin{table}[htbp]
	\centering
	\vspace{-2mm}
	\caption{Quantitative comparison of the proposed method with the “joint training” methods. The number of parameters and the computational cost of the trainable parts are reported in the table. The best and second-best performances for each metric are highlighted with {\setlength{\fboxsep}{2.3pt}\colorbox{lightred}{Red}} and {\setlength{\fboxsep}{2.3pt}\colorbox{lightblue}{Blue}} backgrounds, respectively.}\vspace{-2mm}
	\renewcommand\arraystretch{1.4}
	{\footnotesize\centerline{\tabcolsep=2.5pt
			\begin{tabular}{c|c|c|cc|c|c}
				\toprule
				\multirow{2}{*}{\textbf{Methods}} & \textbf{OD}    & \textbf{Seg}   & \multicolumn{2}{c|}{\textbf{SOD}} & \multicolumn{1}{c|}{\multirow{2}{*}{\thead{\textbf{Params}\\(M)}}} & \multirow{2}{*}{\thead{\textbf{FLOPs}\\(G)}} \\
				\cline{2-5}          & mAP$_{50 \rightarrow 95} \uparrow$   & mIoU $\uparrow$  & mF$_{\beta} \uparrow$    & E$_m \uparrow$    &       &  \\
				\midrule
				IRFS\cite{11}  & \cellcolor{lightblue}{0.6180} & 59.10  & \cellcolor{lightred}{0.8319} & \cellcolor{lightred}{0.9194} & \cellcolor{lightblue}{39.95} & 253.08  \\
				MRFS\cite{7}  & 0.5959  & \cellcolor{lightred}{61.48} & 0.7800  & 0.8879  & 134.97  & \cellcolor{lightblue}{219.16} \\
				SMiF\cite{16} & 0.6092  & \cellcolor{lightblue}{61.20} & 0.8004  & 0.9012  & 45.60  & 526.20  \\
				
				Ours  & \cellcolor{lightred}{0.6304} & 60.34  & \cellcolor{lightblue}{0.8129} & \cellcolor{lightblue}{0.9087} &\cellcolor{lightred}{0.46} & \cellcolor{lightred}{174.06} \\
				\bottomrule
	\end{tabular}}}\vspace{-2.5mm}
	\label{tab:3}%
\end{table}%

\begin{figure}[t!]
	\centering
	\includegraphics[width=0.45\textwidth]{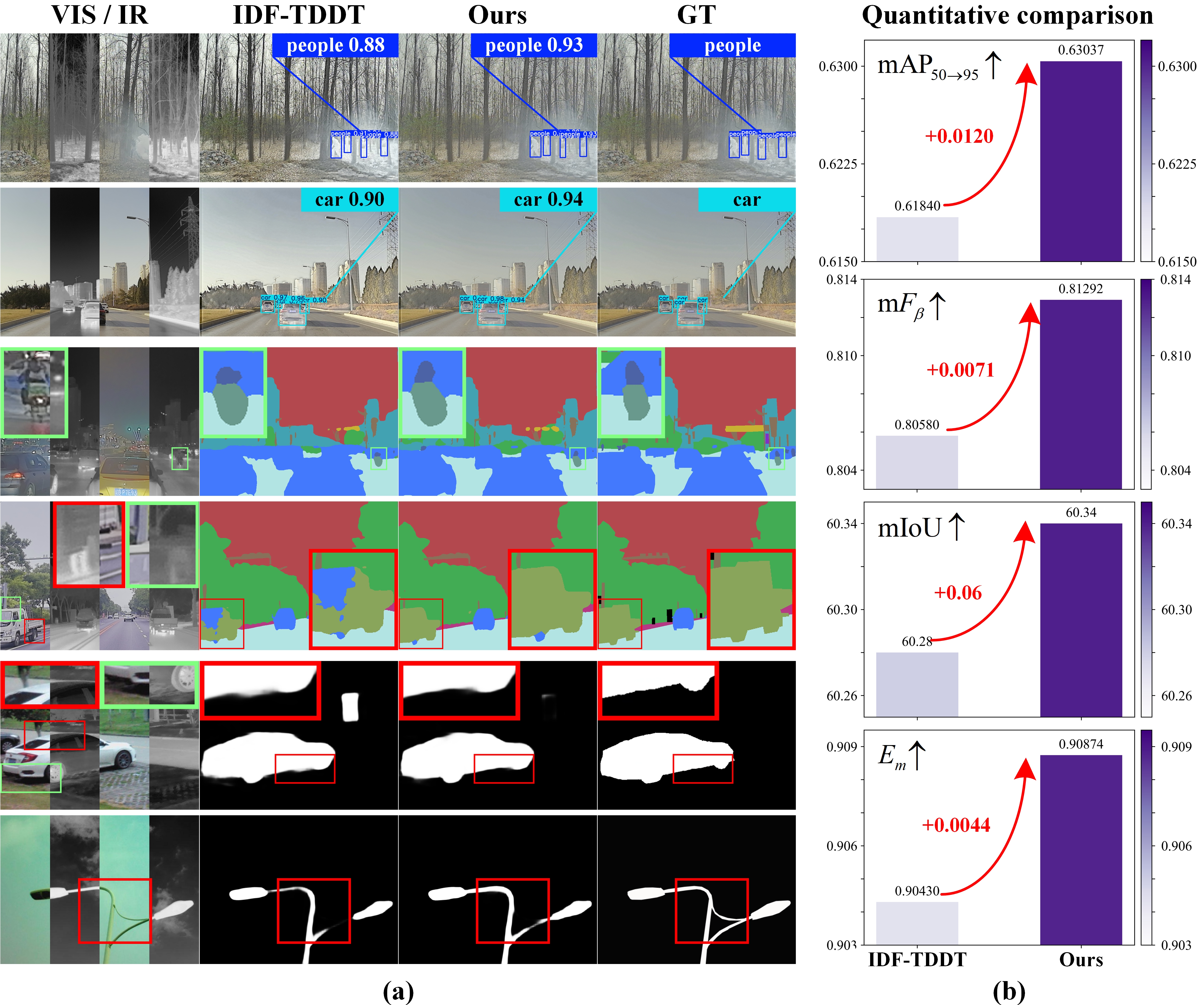}\vspace{-2mm}
	\caption{Qualitative (a) and quantitative (b) comparison between the proposed method and IDF-TDDT.}\vspace{-3mm}
	\label{label7}
\end{figure}

\textbf{Further Analysis}. The latest method IDF-TDDT uses instructions from multiple downstream tasks to fine-tune the fusion network for improved adaptability across diverse tasks Therefore, we compare our method with IDF-TDDT. As shown in Figure \ref{label7}, our method significantly outperforms IDF-TDDT in both quantitative and qualitative evaluations across multiple downstream tasks. In addition, IDF-TDDT encodes task instructions using LLaMA \cite{52}, which introduces substantial parameter and computational overhead, limiting its deployment on resource-constrained platforms.

\begin{figure}[t!]
	\centering
	\includegraphics[width=0.45\textwidth]{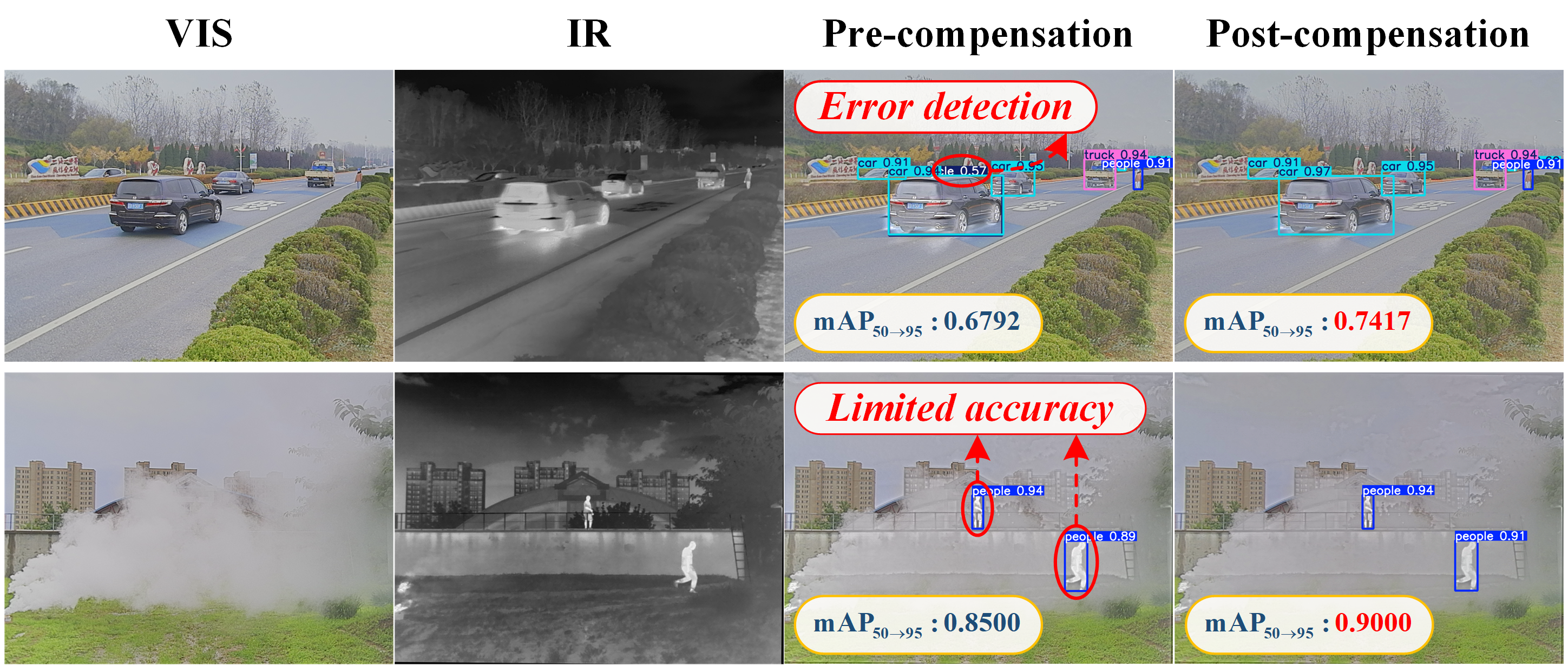}\vspace{-2mm}
	\caption{Qualitative and quantitative comparison between pre-compensation and post-compensation results.}\vspace{-3mm}
	\label{label8}
\end{figure}

\subsection{Cross-Detector Generalization}
To evaluate cross-detector generalization, we replace YOLOv5 with DETR \cite{63} and deploy the semantically compensated fused images to DETR without retraining the RSC module. As shown in Table \ref{tab:cdg}, the compensated results achieve improvements in mAP$_{50 \rightarrow 95}$ over the original VFN results on both detectors. These quantitative results demonstrate that our method remains effective under cross-detector settings.

\begin{table}[htbp]
	\centering
	\caption{Quantitative results of cross-detector generalization.}\vspace{-2mm}
	\renewcommand\arraystretch{1.2}
	{\footnotesize\centerline{\tabcolsep=23pt
			\begin{tabular}{c|c|c}
				\toprule
				\textbf{Detectors} & \multicolumn{1}{c|}{\textbf{VFN}} & \textbf{VFN + RSC} \\
				\midrule
				DETR \cite{63}  & 0.5610  & 0.5810 \\
				YOLOv5 & 0.6076 & 0.6304 \\
				\bottomrule
	\end{tabular}}}\vspace{-3mm}
	\label{tab:cdg}%
\end{table}%

\subsection{Ablation Study}
In our method, the closed-loop optimization mechanism and the RSC module serve as the core components. The penalty loss $\ell_{r}^n$ plays a key role in the closed-loop optimization mechanism. We therefore validate the effectiveness of the closed-loop optimization mechanism, $\ell_{r}^n$, and the RSC on the M$^3$FD, FMB, and VT5000 datasets.

\textbf{Effectiveness of Semantic Compensation}. We compare the task performance before and after compensation. As shown in Figure \ref{label8}, taking the OD as an example, the pre-compensation results suffer from "\emph{Error Detection}" and "\emph{Limited Accuracy}" issues, leading to a lower mAP$_{50 \rightarrow 95}$. In contrast, the post-compensation results effectively resolve these issues, leading to improved mAP$_{50 \rightarrow 95}$.

\begin{table}[htbp]
	\centering
	\caption{Quantitative comparison between the full model and its ablated variants. The best performance for each metric is marked with a {\setlength{\fboxsep}{2.3pt}\colorbox{lightred}{Red}} background.}\vspace{-2mm}
	\renewcommand\arraystretch{1.3}
	{\footnotesize\centerline{\tabcolsep=9pt
			\begin{tabular}{c|c|c|cc}
				\toprule
				\multirow{2}{*}{\textbf{Methods}} & \textbf{OD}    & \textbf{Seg}   & \multicolumn{2}{c}{\textbf{SOD}} \\
				\cline{2-5}          &mAP$_{50 \rightarrow 95} \uparrow$   & mIoU $\uparrow$  & mF$_{\beta} \uparrow$    & E$_m \uparrow$ \\
				\midrule
				Model I & 0.6272  & 60.15 & \cellcolor{lightred}{0.8136} & \cellcolor{lightred}{0.9091} \\
				Model II & 0.6276  & 60.18 & 0.8134  & 0.9091  \\
				Model III & 0.6298  & 60.07 & 0.8115  & 0.9081  \\
				Ours  & \cellcolor{lightred}{0.6304} & \cellcolor{lightred}{60.34} & 0.8129  & 0.9087  \\
				\bottomrule
	\end{tabular}}}\vspace{-3mm}
	\label{tab:4}%
\end{table}%

\begin{figure}[t!]
	\centering
	\includegraphics[width=0.48\textwidth]{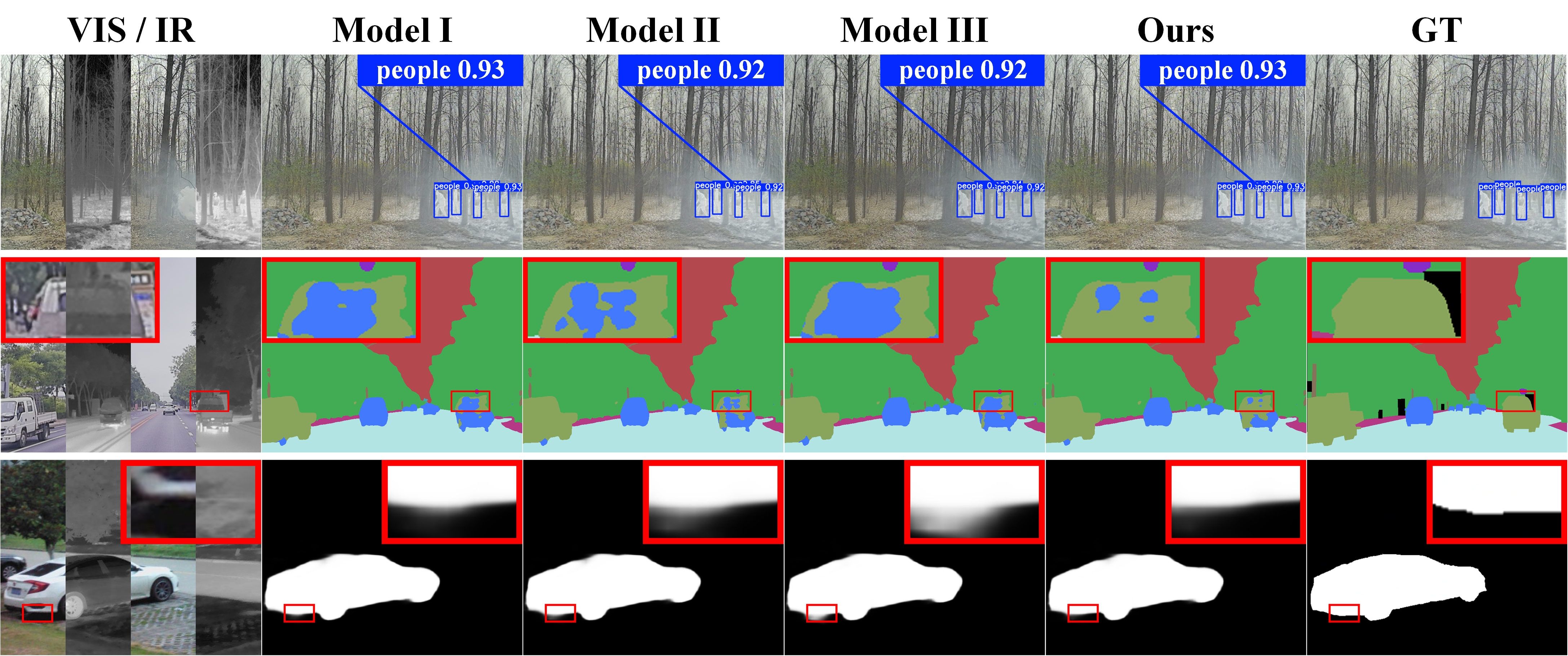}
	\caption{Qualitative comparison between the full model and its ablated variants.}\vspace{-3mm}
	\label{label9}\vspace{-3mm}
\end{figure}

\textbf{Effectiveness of the Closed-Loop Optimization Mechanism}. Model I is designed, in which the fused images are directly fed into the multi-task networks, and the RSC is optimized only with task losses. As shown in Table \ref{tab:4} and Figure \ref{label9}, Model I exhibits a noticeable task bias. In contrast, the full model achieves stable performance across downstream tasks.

\textbf{Effectiveness of the Penalty Loss $\ell_{r}^n$}. Model II is designed such that it removes $\ell_{r}^n$ and optimizes the RSC solely with downstream task losses. As shown in Table \ref{tab:4} and Figure \ref{label9}, Model II exhibits a clear bias toward the salient object detection task. In contrast, the full model achieves more balanced and superior performance across multiple tasks.

\textbf{Effectiveness of RSC}. Model III is designed such that the RSC is replaced with convolutional networks. As shown in Table \ref{tab:4} and Figure \ref{label9}, Model III exhibits inferior multi-task adaptability compared with the full model.

\section{Conclusion}
We propose Closed-Loop Dynamic Network (CLDyN), a novel infrared-visible image fusion framework for multi-task adaptation. CLDyN introduces a closed-loop optimization mechanism that adaptively performs task-specific semantic compensation on the fusion network according to the semantic requirements of various downstream tasks, thereby enabling the fusion network to dynamically adapt to multiple tasks without retraining. Within this mechanism, a Requirement-driven Semantic Compensation (RSC) module is designed, which customizes the network architecture based on task demands to supplement task-specific semantic information. Experiments on object detection, semantic segmentation, and salient object detection demonstrate that CLDyN achieves superior performance and exhibits strong adaptability across multiple tasks.

\section*{Acknowledgments}
This work was supported in part by the National Natural Science
Foundation of China under Grant 62576132, Grant U23A20294, Grant 62571222, and Grant 62276120, and in part by the Fundamental and Interdisciplinary Disciplines Breakthrough Plan of the Ministry of Education of China under Grant JYB2025XDXM109, and in part by the Yunnan Fundamental Research Projects under Grant 202501AS070123 and Grant 202301AV070004.

{
    \small
    \bibliographystyle{ieeenat_fullname}
    \bibliography{main}

\begin{thebibliography}{60}
\providecommand{\natexlab}[1]{#1}
\providecommand{\url}[1]{\texttt{#1}}
\expandafter\ifx\csname urlstyle\endcsname\relax
  \providecommand{\doi}[1]{doi: #1}\else
  \providecommand{\doi}{doi: \begingroup \urlstyle{rm}\Url}\fi

\bibitem[Bai et~al.(2025{\natexlab{a}})Bai, Zhang, Zhao, Wu, Deng, Cui, Feng,
  and Xu]{4}
Haowen Bai, Jiangshe Zhang, Zixiang Zhao, Yichen Wu, Lilun Deng, Yukun Cui, Tao
  Feng, and Shuang Xu.
\newblock Task-driven image fusion with learnable fusion loss.
\newblock In \emph{Proceedings of the IEEE/CVF Conference on Computer Vision
  and Pattern Recognition (CVPR)}, pages 7457--7468, 2025{\natexlab{a}}.

\bibitem[Bai et~al.(2025{\natexlab{b}})Bai, Zhao, Zhang, Jiang, Deng, Cui, Xu,
  and Zhang]{6}
Haowen Bai, Zixiang Zhao, Jiangshe Zhang, Baisong Jiang, Lilun Deng, Yukun Cui,
  Shuang Xu, and Chunxia Zhang.
\newblock Deep unfolding multi-modal image fusion network via attribution
  analysis.
\newblock \emph{IEEE Transactions on Circuits and Systems for Video
  Technology}, 35\penalty0 (4):\penalty0 3498--3511, 2025{\natexlab{b}}.

\bibitem[Bu et~al.(2024)Bu, Zeng, Chen, Yang, Zhou, Yan, Luo, Cui, Ma, and
  Li]{68}
Qingwen Bu, Jia Zeng, Li Chen, Yanchao Yang, Guyue Zhou, Junchi Yan, Ping Luo,
  Heming Cui, Yi Ma, and Hongyang Li.
\newblock Closed-loop visuomotor control with generative expectation for
  robotic manipulation.
\newblock In \emph{Advances in Neural Information Processing Systems
  (NeurIPS)}, pages 139002--139029, 2024.

\bibitem[Cao et~al.(2024)Cao, Xu, Zhu, Wang, and Hu]{24}
Bing Cao, Xingxin Xu, Pengfei Zhu, Qilong Wang, and Qinghua Hu.
\newblock Conditional controllable image fusion.
\newblock In \emph{Advances in Neural Information Processing Systems
  (NeurIPS)}, pages 120311--120335, 2024.

\bibitem[Carion et~al.(2020)Carion, Massa, Synnaeve, Usunier, Kirillov, and
  Zagoruyko]{63}
Nicolas Carion, Francisco Massa, Gabriel Synnaeve, Nicolas Usunier, Alexander
  Kirillov, and Sergey Zagoruyko.
\newblock End-to-end object detection with transformers.
\newblock In \emph{Proceedings of the European Conference on Computer Vision
  (ECCV)}, pages 213--229, 2020.

\bibitem[Chen and Varshney(2007)]{44}
Hao Chen and Pramod~K. Varshney.
\newblock A human perception inspired quality metric for image fusion based on
  regional information.
\newblock \emph{Information Fusion}, 8\penalty0 (2):\penalty0 193--207, 2007.

\bibitem[Chen et~al.(2025)Chen, Yang, Yu, Gong, Cai, and Ma]{19}
Jun Chen, Liling Yang, Wei Yu, Wenping Gong, Zhanchuan Cai, and Jiayi Ma.
\newblock Sdsfusion: A semantic-aware infrared and visible image fusion network
  for degraded scenes.
\newblock \emph{IEEE Transactions on Image Processing}, 34:\penalty0
  3139--3153, 2025.

\bibitem[Chen and Blum(2009)]{43}
Yin Chen and Rick~S. Blum.
\newblock A new automated quality assessment algorithm for image fusion.
\newblock \emph{Image and Vision Computing}, 27\penalty0 (10):\penalty0
  1421--1432, 2009.

\bibitem[Chen et~al.(2020)Chen, Dai, Liu, Chen, Yuan, and Liu]{66}
Yinpeng Chen, Xiyang Dai, Mengchen Liu, Dongdong Chen, Lu Yuan, and Zicheng
  Liu.
\newblock Dynamic convolution: Attention over convolution kernels.
\newblock In \emph{Proceedings of the IEEE/CVF Conference on Computer Vision
  and Pattern Recognition (CVPR)}, pages 11027--11036, 2020.

\bibitem[Cheng et~al.(2025)Cheng, Xu, Feng, Wu, Tang, Li, Zhang, Atito, Awais,
  and Kittler]{25}
Chunyang Cheng, Tianyang Xu, Zhenhua Feng, Xiaojun Wu, Zhangyong Tang, Hui Li,
  Zeyang Zhang, Sara Atito, Muhammad Awais, and Josef Kittler.
\newblock One model for all: Low-level task interaction is a key to
  task-agnostic image fusion.
\newblock In \emph{Proceedings of the IEEE/CVF Conference on Computer Vision
  and Pattern Recognition (CVPR)}, pages 28102--28112, 2025.

\bibitem[Erickson et~al.(2018)Erickson, Clever, Turk, Liu, and Kemp]{69}
Zackory Erickson, Henry~M. Clever, Greg Turk, C.~Karen Liu, and Charles~C.
  Kemp.
\newblock Deep haptic model predictive control for robot-assisted dressing.
\newblock In \emph{2018 IEEE International Conference on Robotics and
  Automation (ICRA)}, pages 4437--4444, 2018.

\bibitem[Guo et~al.(2025)Guo, Luo, Liu, Zhang, and Wu]{20}
Lin Guo, Xiaoqing Luo, Yue Liu, Zhancheng Zhang, and Xiaojun Wu.
\newblock Sam-guided multi-level collaborative transformer for infrared and
  visible image fusion.
\newblock \emph{Pattern Recognition}, 162:\penalty0 111391, 2025.

\bibitem[Han et~al.(2022)Han, Huang, Song, Yang, Wang, and Wang]{67}
Yizeng Han, Gao Huang, Shiji Song, Le Yang, Honghui Wang, and Yulin Wang.
\newblock Dynamic neural networks: A survey.
\newblock \emph{IEEE Transactions on Pattern Analysis and Machine
  Intelligence}, 44\penalty0 (11):\penalty0 7436--7456, 2022.

\bibitem[Jia et~al.(2021)Jia, Zhu, Li, Tang, and Zhou]{33}
Xinyu Jia, Chuang Zhu, Minzhen Li, Wenqi Tang, and Wenli Zhou.
\newblock Llvip: A visible-infrared paired dataset for low-light vision.
\newblock In \emph{Proceedings of the IEEE/CVF International Conference on
  Computer Vision Workshops (ICCVW)}, pages 3496--3504, 2021.

\bibitem[Kingma and Ba(2015)]{39}
Diederik~P. Kingma and Jimmy Ba.
\newblock Adam: A method for stochastic optimization.
\newblock In \emph{Proceedings of International Conference on Learning
  Representations (ICLR)}, 2015.

\bibitem[Kirillov et~al.(2023)Kirillov, Mintun, Ravi, Mao, Rolland, Gustafson,
  Xiao, Whitehead, Berg, Lo, Dollar, and Girshick]{31}
Alexander Kirillov, Eric Mintun, Nikhila Ravi, Hanzi Mao, Chloe Rolland, Laura
  Gustafson, Tete Xiao, Spencer Whitehead, Alexander~C. Berg, Wan-Yen Lo, Piotr
  Dollar, and Ross Girshick.
\newblock Segment anything.
\newblock In \emph{Proceedings of the IEEE/CVF International Conference on
  Computer Vision (ICCV)}, pages 4015--4026, 2023.

\bibitem[Li et~al.(2025{\natexlab{a}})Li, Bian, Zhang, Song, Li, and Wu]{62}
Hui Li, Congcong Bian, Zeyang Zhang, Xiaoning Song, Xi Li, and XiaoJun Wu.
\newblock Occo: Lvm-guided infrared and visible image fusion framework based on
  object-aware and contextual contrastive learning.
\newblock \emph{International Journal of Computer Vision}, 133\penalty0
  (9):\penalty0 6611--6635, 2025{\natexlab{a}}.

\bibitem[Li et~al.(2025{\natexlab{b}})Li, Bian, Zhang, Song, Li, and Wu]{30}
Hui Li, Congcong Bian, Zeyang Zhang, Xiaoning Song, Xi Li, and Xiao-Jun Wu.
\newblock Occo: Lvm-guided infrared and visible image fusion framework based on
  object-aware and contextual contrastive learning.
\newblock \emph{International Journal of Computer Vision}, 133\penalty0
  (9):\penalty0 6611--6635, 2025{\natexlab{b}}.

\bibitem[Li et~al.(2025{\natexlab{c}})Li, Yang, Zhang, Jia, Yu, and Liu]{21}
Huafeng Li, Zengyi Yang, Yafei Zhang, Wei Jia, Zhengtao Yu, and Yu Liu.
\newblock Mulfs-cap: Multimodal fusion-supervised cross-modality alignment
  perception for unregistered infrared-visible image fusion.
\newblock \emph{IEEE Transactions on Pattern Analysis and Machine
  Intelligence}, 47\penalty0 (5):\penalty0 3673--3690, 2025{\natexlab{c}}.

\bibitem[Li et~al.(2023)Li, Zou, Liu, Jiang, Ma, Fan, and Liu]{58}
Xingyuan Li, Yang Zou, Jinyuan Liu, Zhiying Jiang, Long Ma, Xin Fan, and
  Risheng Liu.
\newblock From text to pixels: A context-aware semantic synergy solution for
  infrared and visible image fusion.
\newblock \emph{arXiv preprint arXiv:2401.00421}, 2023.

\bibitem[Li et~al.(2024)Li, Liu, Chen, Zou, Ma, Fan, and Liu]{56}
Xingyuan Li, Jinyuan Liu, Zhixin Chen, Yang Zou, Long Ma, Xin Fan, and Risheng
  Liu.
\newblock Contourlet residual for prompt learning enhanced infrared image
  super-resolution.
\newblock In \emph{European Conference on Computer Vision}, pages 270--288.
  Springer, 2024.

\bibitem[Li et~al.(2025{\natexlab{d}})Li, Wang, Zou, Chen, Ma, Jiang, Ma, and
  Liu]{57}
Xingyuan Li, Zirui Wang, Yang Zou, Zhixin Chen, Jun Ma, Zhiying Jiang, Long Ma,
  and Jinyuan Liu.
\newblock Difiisr: A diffusion model with gradient guidance for infrared image
  super-resolution.
\newblock In \emph{Proceedings of the Computer Vision and Pattern Recognition
  Conference}, pages 7534--7544, 2025{\natexlab{d}}.

\bibitem[Liang et~al.(2024)Liang, Jiang, Ma, Liu, and Ma]{18}
Pengwei Liang, Junjun Jiang, Qing Ma, Xianming Liu, and Jiayi Ma.
\newblock Fusion from decomposition: A self-supervised approach for image
  fusion and beyond.
\newblock \emph{arXiv preprint arXiv: 2410.12274}, 2024.

\bibitem[Liu et~al.(2021)Liu, Liu, Jin, Stone, and Liu]{64}
Bo Liu, Xingchao Liu, Xiaojie Jin, Peter Stone, and Qiang Liu.
\newblock Conflict-averse gradient descent for multi-task learning.
\newblock In \emph{Advances in Neural Information Processing Systems
  (NeurIPS)}, pages 18878--18890, 2021.

\bibitem[Liu et~al.(2022)Liu, Fan, Huang, Wu, Liu, Zhong, and Luo]{10}
Jinyuan Liu, Xin Fan, Zhanbo Huang, Guanyao Wu, Risheng Liu, Wei Zhong, and
  Zhongxuan Luo.
\newblock Target-aware dual adversarial learning and a multi-scenario
  multi-modality benchmark to fuse infrared and visible for object detection.
\newblock In \emph{Proceedings of the IEEE/CVF Conference on Computer Vision
  and Pattern Recognition (CVPR)}, pages 5802--5811, 2022.

\bibitem[Liu et~al.(2023{\natexlab{a}})Liu, Liu, Wu, Ma, Liu, Zhong, Luo, and
  Fan]{16}
Jinyuan Liu, Zhu Liu, Guanyao Wu, Long Ma, Risheng Liu, Wei Zhong, Zhongxuan
  Luo, and Xin Fan.
\newblock Multi-interactive feature learning and a full-time multi-modality
  benchmark for image fusion and segmentation.
\newblock In \emph{Proceedings of the IEEE/CVF International Conference on
  Computer Vision (ICCV)}, pages 8081--8090, 2023{\natexlab{a}}.

\bibitem[Liu et~al.(2024{\natexlab{a}})Liu, Lin, Wu, Liu, Luo, and Fan]{51}
Jinyuan Liu, Runjia Lin, Guanyao Wu, Risheng Liu, Zhongxuan Luo, and Xin Fan.
\newblock Coconet: Coupled contrastive learning network with multi-level
  feature ensemble for multi-modality image fusion.
\newblock \emph{International Journal of Computer Vision}, 132\penalty0
  (5):\penalty0 1748--1775, 2024{\natexlab{a}}.

\bibitem[Liu et~al.(2025{\natexlab{a}})Liu, Wu, Liu, Wang, Jiang, Ma, Zhong,
  Fan, and Liu]{45}
Jinyuan Liu, Guanyao Wu, Zhu Liu, Di Wang, Zhiying Jiang, Long Ma, Wei Zhong,
  Xin Fan, and Risheng Liu.
\newblock Infrared and visible image fusion: From data compatibility to task
  adaption.
\newblock \emph{IEEE Transactions on Pattern Analysis and Machine
  Intelligence}, 47\penalty0 (4):\penalty0 2349--2369, 2025{\natexlab{a}}.

\bibitem[Liu et~al.(2025{\natexlab{b}})Liu, Wu, Liu, Wang, Jiang, Ma, Zhong,
  Fan, and Liu]{60}
Jinyuan Liu, Guanyao Wu, Zhu Liu, Di Wang, Zhiying Jiang, Long Ma, Wei Zhong,
  Xin Fan, and Risheng Liu.
\newblock Infrared and visible image fusion: From data compatibility to task
  adaption.
\newblock \emph{IEEE Transactions on Pattern Analysis and Machine
  Intelligence}, 47\penalty0 (4):\penalty0 2349--2369, 2025{\natexlab{b}}.

\bibitem[Liu et~al.(2024{\natexlab{b}})Liu, Liu, Liu, Fan, and Luo]{14}
Risheng Liu, Zhu Liu, Jinyuan Liu, Xin Fan, and Zhongxuan Luo.
\newblock A task-guided, implicitly-searched and meta-initialized deep model
  for image fusion.
\newblock \emph{IEEE Transactions on Pattern Analysis and Machine
  Intelligence}, 46\penalty0 (10):\penalty0 6594--6609, 2024{\natexlab{b}}.

\bibitem[Liu et~al.(2024{\natexlab{c}})Liu, Qi, Cheng, and Chen]{53}
Yu Liu, Zhengzheng Qi, Juan Cheng, and Xun Chen.
\newblock Rethinking the effectiveness of objective evaluation metrics in
  multi-focus image fusion: A statistic-based approach.
\newblock \emph{IEEE Transactions on Pattern Analysis and Machine
  Intelligence}, 46\penalty0 (8):\penalty0 5806--5819, 2024{\natexlab{c}}.

\bibitem[Liu et~al.(2023{\natexlab{b}})Liu, Liu, Wu, Ma, Fan, and Liu]{13}
Zhu Liu, Jinyuan Liu, Guanyao Wu, Long Ma, Xin Fan, and Risheng Liu.
\newblock Bi-level dynamic learning for jointly multi-modality image fusion and
  beyond.
\newblock In \emph{Proceedings of the Thirty-Second International Joint
  Conference on Artificial Intelligence (IJCAI)}, pages 1240--1248,
  2023{\natexlab{b}}.

\bibitem[Liu et~al.(2023{\natexlab{c}})Liu, Liu, Zhang, Ma, Fan, and Liu]{12}
Zhu Liu, Jinyuan Liu, Benzhuang Zhang, Long Ma, Xin Fan, and Risheng Liu.
\newblock Paif: Perception-aware infrared-visible image fusion for
  attack-tolerant semantic segmentation.
\newblock In \emph{Proceedings of the 31st ACM International Conference on
  Multimedia (ACM MM)}, page 3706–3714, 2023{\natexlab{c}}.

\bibitem[Ma et~al.(2019)Ma, Ma, and Li]{42}
Jiayi Ma, Yong Ma, and Chang Li.
\newblock Infrared and visible image fusion methods and applications: A survey.
\newblock \emph{Information Fusion}, 45:\penalty0 153--178, 2019.

\bibitem[Shi et~al.(2025)Shi, Liu, Cheng, Wang, and Chen]{54}
Yu Shi, Yu Liu, Juan Cheng, Z.~Jane Wang, and Xun Chen.
\newblock Vdmufusion: A versatile diffusion model-based unsupervised framework
  for image fusion.
\newblock \emph{IEEE Transactions on Image Processing}, 34:\penalty0 441--454,
  2025.

\bibitem[Sun et~al.(2022)Sun, Cao, Zhu, and Hu]{5}
Yiming Sun, Bing Cao, Pengfei Zhu, and Qinghua Hu.
\newblock Detfusion: A detection-driven infrared and visible image fusion
  network.
\newblock In \emph{Proceedings of the 30th ACM International Conference on
  Multimedia (ACM MM)}, page 4003–4011, 2022.

\bibitem[Sun et~al.(2025)Sun, Li, Zhu, Hu, Ren, Xu, and Zhu]{29}
Yiming Sun, Xin Li, Pengfei Zhu, Qinghua Hu, Dongwei Ren, Huiying Xu, and
  Xinzhong Zhu.
\newblock Task-gated multi-expert collaboration network for degraded
  multi-modal image fusion.
\newblock In \emph{Proceedings of 42nd International Conference on Machine
  Learning (ICML)}, 2025.

\bibitem[Tang et~al.(2022{\natexlab{a}})Tang, Yuan, and Ma]{1}
Linfeng Tang, Jiteng Yuan, and Jiayi Ma.
\newblock Image fusion in the loop of high-level vision tasks: A semantic-aware
  real-time infrared and visible image fusion network.
\newblock \emph{Information Fusion}, 82:\penalty0 28--42, 2022{\natexlab{a}}.

\bibitem[Tang et~al.(2022{\natexlab{b}})Tang, Yuan, Zhang, Jiang, and Ma]{34}
Linfeng Tang, Jiteng Yuan, Hao Zhang, Xingyu Jiang, and Jiayi Ma.
\newblock Piafusion: A progressive infrared and visible image fusion network
  based on illumination aware.
\newblock \emph{Information Fusion}, 83-84:\penalty0 79--92,
  2022{\natexlab{b}}.

\bibitem[Tang et~al.(2023)Tang, Zhang, Xu, and Ma]{17}
Linfeng Tang, Hao Zhang, Han Xu, and Jiayi Ma.
\newblock Rethinking the necessity of image fusion in high-level vision tasks:
  A practical infrared and visible image fusion network based on progressive
  semantic injection and scene fidelity.
\newblock \emph{Information Fusion}, 99:\penalty0 101870, 2023.

\bibitem[Tang et~al.(2025)Tang, Yan, Xiang, Fang, and Ma]{23}
Linfeng Tang, Qinglong Yan, Xinyu Xiang, Leyuan Fang, and Jiayi Ma.
\newblock C2rf: Bridging multi-modal image registration and fusion via
  commonality mining and contrastive learning.
\newblock \emph{International Journal of Computer Vision}, 133\penalty0
  (8):\penalty0 5262--5280, 2025.

\bibitem[Touvron et~al.(2023)Touvron, Lavril, Izacard, Martinet, Lachaux,
  Lacroix, Rozière, Goyal, Hambro, Azhar, Rodriguez, Joulin, Grave, and
  Lample]{52}
Hugo Touvron, Thibaut Lavril, Gautier Izacard, Xavier Martinet, Marie-Anne
  Lachaux, Timothée Lacroix, Baptiste Rozière, Naman Goyal, Eric Hambro,
  Faisal Azhar, Aurelien Rodriguez, Armand Joulin, Edouard Grave, and Guillaume
  Lample.
\newblock Llama: Open and efficient foundation language models.
\newblock \emph{arXiv preprint arXiv: 2302.13971}, 2023.

\bibitem[Tu et~al.(2023)Tu, Ma, Li, Li, Xu, and Liu]{32}
Zhengzheng Tu, Yan Ma, Zhun Li, Chenglong Li, Jieming Xu, and Yongtao Liu.
\newblock Rgbt salient object detection: A large-scale dataset and benchmark.
\newblock \emph{IEEE Transactions on Multimedia}, 25:\penalty0 4163--4176,
  2023.

\bibitem[Wang et~al.(2023)Wang, Liu, Liu, and Fan]{11}
Di Wang, Jinyuan Liu, Risheng Liu, and Xin Fan.
\newblock An interactively reinforced paradigm for joint infrared-visible image
  fusion and saliency object detection.
\newblock \emph{Information Fusion}, 98:\penalty0 101828, 2023.

\bibitem[Wang et~al.(2025)Wang, Jiao, Liu, and Fan]{22}
Di Wang, Xianghao Jiao, Jinyuan Liu, and Xin Fan.
\newblock Robust one-stop multi-modality image registration-fusion-segmentation
  framework against misalignments and adversarial attacks.
\newblock \emph{IEEE Transactions on Multimedia}, 27:\penalty0 4531--4543,
  2025.

\bibitem[Wu et~al.(2025)Wu, Liu, Fu, Peng, Liu, Fan, and Liu]{8}
Guanyao Wu, Haoyu Liu, Hongming Fu, Yichuan Peng, Jinyuan Liu, Xin Fan, and
  Risheng Liu.
\newblock Every sam drop counts: Embracing semantic priors for multi-modality
  image fusion and beyond.
\newblock In \emph{Proceedings of the IEEE/CVF Conference on Computer Vision
  and Pattern Recognition (CVPR)}, pages 17882--17891, 2025.

\bibitem[Xie et~al.(2021)Xie, Wang, Yu, Anandkumar, Alvarez, and Luo]{37}
Enze Xie, Wenhai Wang, Zhiding Yu, Anima Anandkumar, Jose~M Alvarez, and Ping
  Luo.
\newblock Segformer: Simple and efficient design for semantic segmentation with
  transformers.
\newblock In \emph{Advances in Neural Information Processing Systems
  (NeurIPS)}, 2021.

\bibitem[Xu et~al.(2020)Xu, Ma, Le, Jiang, and Guo]{36}
Han Xu, Jiayi Ma, Zhuliang Le, Junjun Jiang, and Xiaojie Guo.
\newblock Fusiondn: A unified densely connected network for image fusion.
\newblock \emph{Proceedings of the AAAI Conference on Artificial Intelligence},
  34\penalty0 (07):\penalty0 12484--12491, 2020.

\bibitem[Xu et~al.(2022)Xu, Ma, Jiang, Guo, and Ling]{35}
Han Xu, Jiayi Ma, Junjun Jiang, Xiaojie Guo, and Haibin Ling.
\newblock U2fusion: A unified unsupervised image fusion network.
\newblock \emph{IEEE Transactions on Pattern Analysis and Machine
  Intelligence}, 44\penalty0 (1):\penalty0 502--518, 2022.

\bibitem[Xydeas et~al.(2000)Xydeas, Petrovic, et~al.]{41}
Costas~S Xydeas, Vladimir Petrovic, et~al.
\newblock Objective image fusion performance measure.
\newblock \emph{Electronics letters}, 36\penalty0 (4):\penalty0 308--309, 2000.

\bibitem[Yang et~al.(2025)Yang, Zhang, Li, and Liu]{9}
Zengyi Yang, Yafei Zhang, Huafeng Li, and Yu Liu.
\newblock Instruction-driven fusion of infrared–visible images: Tailoring for
  diverse downstream tasks.
\newblock \emph{Information Fusion}, 121:\penalty0 103148, 2025.

\bibitem[Yi et~al.(2024)Yi, Xu, Zhang, Tang, and Ma]{61}
Xunpeng Yi, Han Xu, Hao Zhang, Linfeng Tang, and Jiayi Ma.
\newblock Text-if: Leveraging semantic text guidance for degradation-aware and
  interactive image fusion.
\newblock In \emph{Proceedings of the IEEE/CVF Conference on Computer Vision
  and Pattern Recognition (CVPR)}, pages 27016--27025, 2024.

\bibitem[Zhang et~al.(2024)Zhang, Zuo, Jiang, Guo, and Ma]{7}
Hao Zhang, Xuhui Zuo, Jie Jiang, Chunchao Guo, and Jiayi Ma.
\newblock Mrfs: Mutually reinforcing image fusion and segmentation.
\newblock In \emph{Proceedings of the IEEE/CVF Conference on Computer Vision
  and Pattern Recognition (CVPR)}, pages 26964--26973, 2024.

\bibitem[Zhang et~al.(2025)Zhang, Cao, Zuo, Shao, and Ma]{27}
Hao Zhang, Lei Cao, Xuhui Zuo, Zhenfeng Shao, and Jiayi Ma.
\newblock Omnifuse: Composite degradation-robust image fusion with
  language-driven semantics.
\newblock \emph{IEEE Transactions on Pattern Analysis and Machine
  Intelligence}, 47\penalty0 (9):\penalty0 7577--7595, 2025.

\bibitem[Zhang and Demiris(2023)]{59}
Xingchen Zhang and Yiannis Demiris.
\newblock Visible and infrared image fusion using deep learning.
\newblock \emph{IEEE Transactions on Pattern Analysis and Machine
  Intelligence}, 45\penalty0 (8):\penalty0 10535--10554, 2023.

\bibitem[Zhang et~al.(2020)Zhang, Ye, and Xiao]{40}
Xingchen Zhang, Ping Ye, and Gang Xiao.
\newblock Vifb: A visible and infrared image fusion benchmark.
\newblock In \emph{Proceedings of the IEEE/CVF Conference on Computer Vision
  and Pattern Recognition Workshops (CVPRW)}, pages 468--478, 2020.

\bibitem[Zhao et~al.(2023)Zhao, Xie, Zhao, He, and Lu]{2}
Wenda Zhao, Shigeng Xie, Fan Zhao, You He, and Huchuan Lu.
\newblock Metafusion: Infrared and visible image fusion via meta-feature
  embedding from object detection.
\newblock In \emph{Proceedings of the IEEE/CVF Conference on Computer Vision
  and Pattern Recognition (CVPR)}, pages 13955--13965, 2023.

\bibitem[Zhao et~al.(2025)Zhao, Cui, Wang, He, and Lu]{26}
Wenda Zhao, Hengshuai Cui, Haipeng Wang, You He, and Huchuan Lu.
\newblock Freefusion: Infrared and visible image fusion via cross
  reconstruction learning.
\newblock \emph{IEEE Transactions on Pattern Analysis and Machine
  Intelligence}, 47\penalty0 (9):\penalty0 8040--8056, 2025.

\bibitem[Zhao et~al.(2021)Zhao, Xia, Xie, and Li]{38}
Zhirui Zhao, Changqun Xia, Chenxi Xie, and Jia Li.
\newblock Complementary trilateral decoder for fast and accurate salient object
  detection.
\newblock In \emph{Proceedings of the 29th ACM International Conference on
  Multimedia (ACM MM)}, page 4967–4975, 2021.

\bibitem[Zhao et~al.(2024)Zhao, Bai, Zhang, Zhang, Zhang, Xu, Chen, Timofte,
  and Van~Gool]{55}
Zixiang Zhao, Haowen Bai, Jiangshe Zhang, Yulun Zhang, Kai Zhang, Shuang Xu,
  Dongdong Chen, Radu Timofte, and Luc Van~Gool.
\newblock Equivariant multi-modality image fusion.
\newblock In \emph{Proceedings of the IEEE/CVF Conference on Computer Vision
  and Pattern Recognition (CVPR)}, pages 25912--25921, 2024.

\end{thebibliography}
}

\clearpage
\appendix

 \clearpage
\setcounter{page}{1}
\maketitlesupplementary

\section{More Details of VFN}
\label{sec:s1}

\begin{figure*}[t!]
	\centering
	\includegraphics[width=0.72\textwidth]{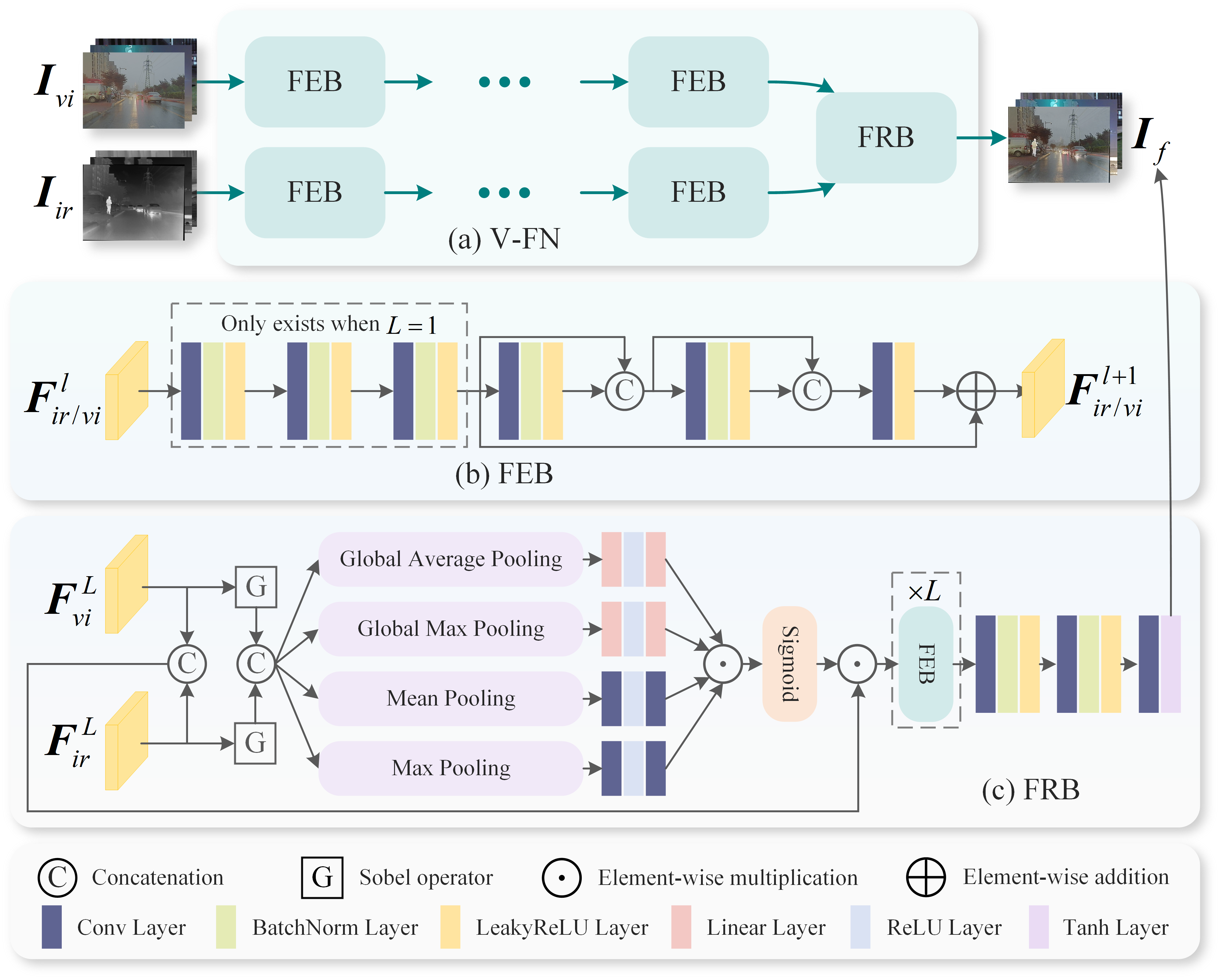}
	\caption{Network architecture of VFN. The VFN (a) consists of a Feature Extraction Blocks (FEB) (b) and a Fusion Feature Reconstruction Block (FRB) (c).}
	\label{labels1}
\end{figure*}

In the first stage, we train the VFN to focus on generating visually guided fused images. In the second stage, the VFN is frozen, while the RSC module assists in adapting the VFN to various downstream task requirements. As shown in Figure \ref{labels1} (a), the VFN consists of 2$L$ Feature Extraction Blocks (FEB) and one Fusion Feature Reconstruction Block (FRB). Each FEB mainly extracts features from infrared or visible images. As illustrated in Figure \ref{labels1} (b), when $L=1$, the FEB contains a single convolutional block designed to enhance the feature channels. When $L>1$, each FEB is composed of three convolutional layers. The FRB is primarily responsible for fusing the extracted infrared and visible features to generate the final fused image. As depicted in Figure \ref{labels1} (c), the FRB consists of Global Average Pooling (GAP), Global Max Pooling (GMP), Mean Pooling (MeanP), Max Pooling (MaxP), and $L$ FEBs.

Specifically, the infrared and visible images ${{\bm I}_{ir/vi}}$ are separately fed into $L$ FEBs to extract features $\{ {\bm F}_{ir/vi}^l\} _{l = 1}^{L - 1}$ from each modality, where ${\bm F}_{ir/vi}^l$ denotes the feature maps obtained from the $l$-th FEB. To reconstruct the fused image, the output feature ${\bm F}_{ir/vi}^L$ from the final FEB is forwarded to the FRB. Within the FRB, we apply the Sobel operator to compute the gradient maps, which are concatenated along the channel dimension to form a combined gradient map ${\bm G} = [\nabla {\bm F}_{ir}^L,\nabla {\bm F}_{vi}^L]$. Here, $[ \cdot ]$ denotes channel-wise concatenation, and $\nabla $ represents the Sobel gradient operator. To enhance the representation of multimodal features, we apply GAP, GMP, MeanP, and MaxP operations on ${\bm G}$, and then combine their outputs through element-wise multiplication. The resulting feature map is normalized by a Sigmoid function to produce the attention matrix ${\bm{A}}$:
\begin{equation}
	\begin{aligned}
		&{{\bm{G}}_f} = {p_f}({\bm{G}}),f \in \{\rm{GAP,GMP,MeanP,MaxP}\} ,\\
		&{\bm{A}} = {\rm{Sigmoid}}(\bm{G}_{GAP} \odot {{\bm{G}}_{GMP}} \odot 	{{\bm{G}}_{MeanP}} \odot {{\bm{G}}_{MaxP}}),
	\end{aligned}
	\label{eq:s1}
\end{equation}
where ${p_f}( \cdot )$ denotes different types of pooling operations, and $ \odot $ represents element-wise multiplication. The attention map ${\bm{A}}$ is then multiplied element-wise with $[{\bm{F}}_{ir}^L,{\bm{F}}_{vi}^L]$, and the result is fed into $L$ FEBs and three convolutional blocks to generate the final fused image ${{\bm{I}}_f}$.

\begin{table}[htbp]
	\centering
	\caption{Quantitative evaluation results on the YOLOv5 and YOLOv11 detectors. The best performance for each metric is marked with a {\setlength{\fboxsep}{2.3pt}\colorbox{lightred}{Red}} background.}
	\renewcommand\arraystretch{1.4}
	{\footnotesize\centerline{\tabcolsep=22pt
	\begin{tabular}{c|c|c}
		\toprule
		\textbf{Methods} & \multicolumn{1}{c|}{\textbf{YOLOv5}} & \multicolumn{1}{c}{\textbf{YOLOv11}} \\
		\midrule
		MRFS \cite{7}  & 0.6189  & 0.6206  \\
		Ours  & \cellcolor{lightred}{0.6304} & \cellcolor{lightred}{0.6430} \\
		\bottomrule
	\end{tabular}}}
	\label{tab:mcad}%
\end{table}%

\begin{table}[htbp]
	\centering
	\caption{Quantitative comparison between the full model and its ablated variants across multiple downstream tasks. The best performance for each metric is marked with a {\setlength{\fboxsep}{2.3pt}\colorbox{lightred}{Red}} background.}\vspace{-2mm}
	\renewcommand\arraystretch{1.4}
	{\footnotesize\centerline{\tabcolsep=8.8pt
			\begin{tabular}{c|c|c|cc}
				\toprule
				\multirow{2}{*}{\textbf{Methods}} & \textbf{OD}    & \textbf{Seg}   & \multicolumn{2}{c}{\textbf{SOD}} \\
				\cline{2-5}          &mAP$_{50 \rightarrow 95} \uparrow$   & mIoU $\uparrow$  & mF$_{\beta} \uparrow$    & E$_m \uparrow$ \\
				\midrule
				Model VI & 0.6289  & 60.13 & 0.8121  & 0.9085  \\
				Model V & 0.6277  & 60.06 & 0.8128  & 0.9086  \\
				Ours  & \cellcolor{lightred}{0.6304} & \cellcolor{lightred}{60.34} & \cellcolor{lightred}{0.8129}  & \cellcolor{lightred}{0.9087}  \\
				\bottomrule
	\end{tabular}}}\vspace{-2mm}
	\label{tab:as_1}%
\end{table}%

\begin{figure*}[t!]
	\centering
	\includegraphics[width=0.95\textwidth]{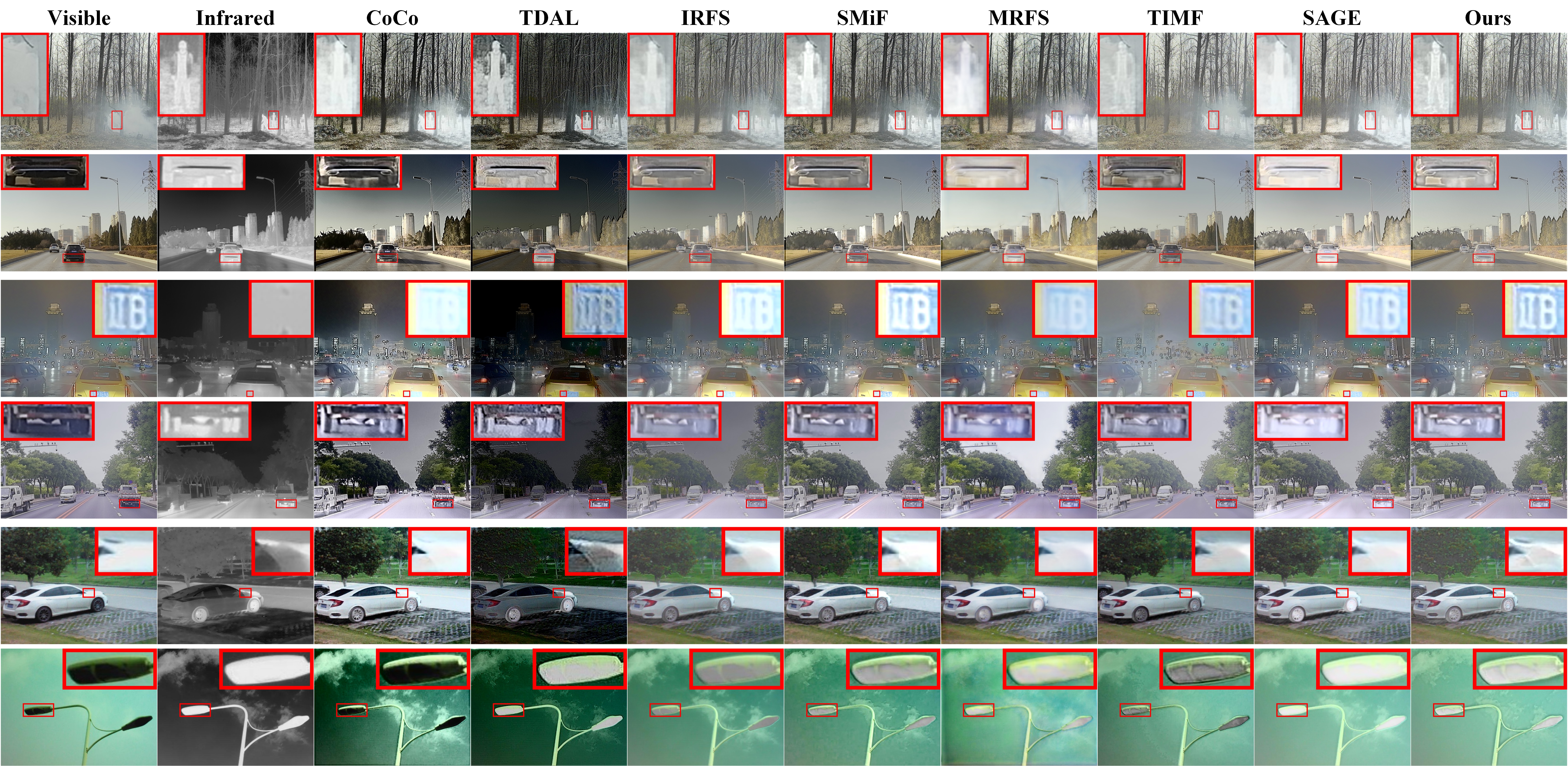}
	\caption{Qualitative comparison between the proposed method and existing state-of-the-art approaches. The first and second columns show the visible and infrared source images, respectively, while the third to tenth columns display the fused results of different comparison methods. The top two rows, middle two rows, and bottom two rows correspond to samples from the M$^3$FD, FMB, and VT5000 datasets, respectively.}
	\label{labels2}
\end{figure*}

\begin{figure}[t!]
	\centering
	\includegraphics[width=0.48\textwidth]{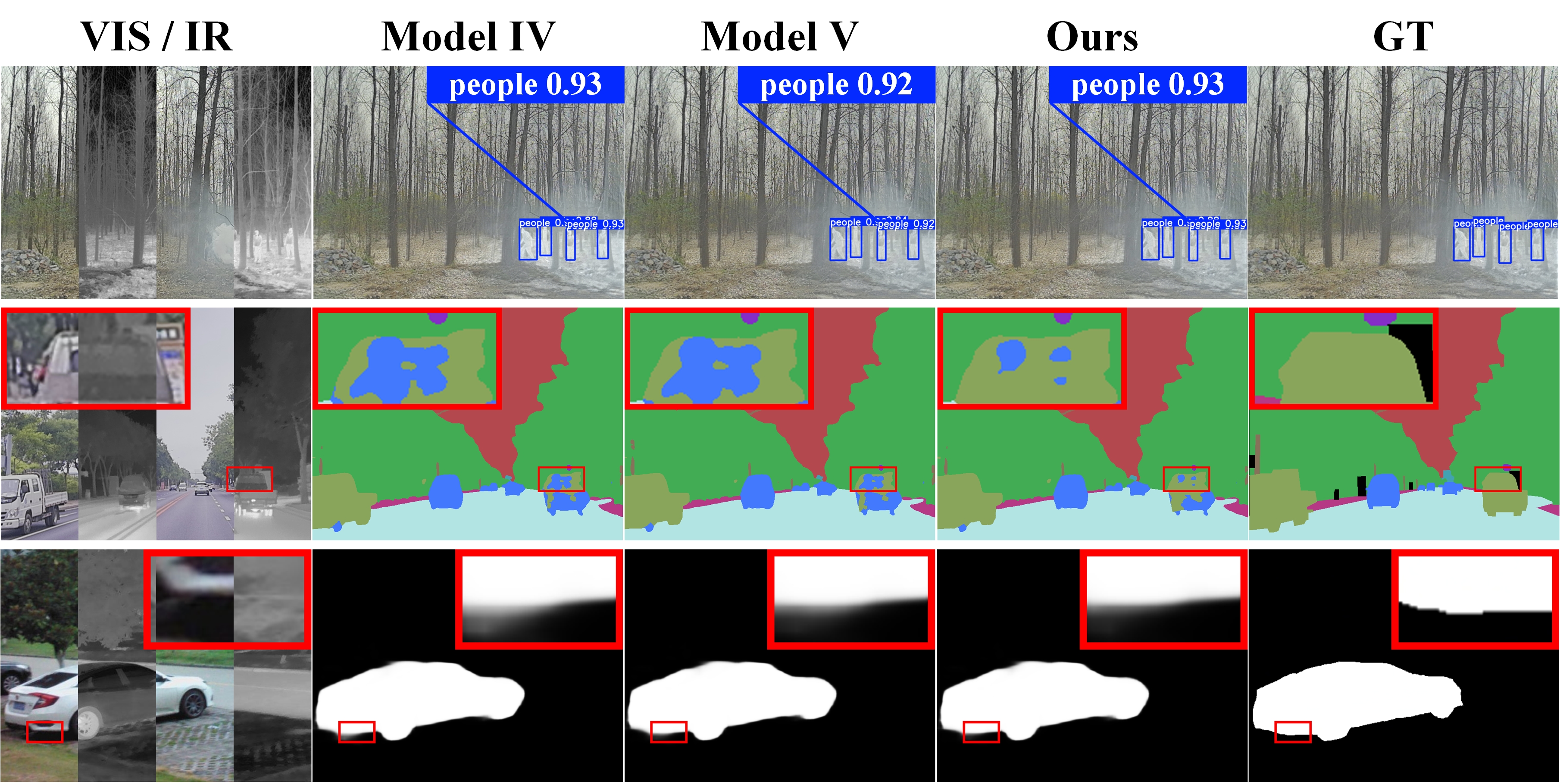}
	\caption{Qualitative comparison between the full model and the ablation models (Model IV and Model V).}
	\label{labels3}
\end{figure}

\begin{table*}[htbp!]
	\centering
	\caption{Quantitative analysis of the hyperparameter $L$ across multiple downstream tasks.  The hyperparameter settings of our proposed method are highlighted in \textbf{bold}, while the best performance for each metric is marked with a {\setlength{\fboxsep}{2.3pt}\colorbox{lightred}{Red}} background.}
	\renewcommand\arraystretch{1.4}
	{\footnotesize\centerline{\tabcolsep=9pt
			\begin{tabular}{c|c|c|cc|c|c}
				\toprule
				\multirow{2}{*}{$L$} & \textbf{OD}    & \textbf{Seg}   & \multicolumn{2}{c|}{\textbf{SOD}} & \multirow{2}{*}{\textbf{Params} (M)} & \multirow{2}{*}{\textbf{FLOPs} (G)} \\
				\cline{2-5}          & mAP$_{50 \rightarrow 95} \uparrow$   & mIoU $\uparrow$  & mF$_{\beta} \uparrow$   & E$_m \uparrow$    &       &  \\
				\midrule
				\textbf{2} & \cellcolor{lightred}{0.6304} & 60.34  & \cellcolor{lightred}{0.8129}  & 0.9087  & \cellcolor{lightred}{0.46} & \cellcolor{lightred}{174.06} \\
				3     & 0.6262  & \cellcolor{lightred}{60.38} & 0.8118   & \cellcolor{lightred}{0.9095} & 0.89  & 348.12 \\
				4     & 0.6300  & 60.55 & 0.8119   & 0.9091  & 1.32  & 522.16 \\
				\bottomrule
	\end{tabular}}}
	\label{tab:ha_L}%
\end{table*}%

\section{More Comparisons on Fusion Performance}
\label{sec:s2}
We further conduct a qualitative comparison with several state-of-the-art fusion methods, including CoCo \cite{51}, TDAL \cite{10}, IRFS \cite{11}, SMiF \cite{16}, MRFS \cite{7}, TIMF \cite{14}, and SAGE \cite{8}. As shown in Figure \ref{labels2}, the qualitative results demonstrate that our method preserves fine details of pedestrians and vehicles while effectively highlighting thermal radiation cues. It achieves a good balance between infrared and visible modalities, producing visually pleasing fusion results. In contrast, existing methods often exhibit modality imbalance, overly emphasizing either infrared or visible information.

\section{More Comparisons on an Advanced Detector}
To further evaluate the adaptability of the proposed method to more advanced detectors, we deploy our method on YOLOv11 for quantitative evaluation. As shown in Table \ref{tab:mcad}, the proposed method achieves significantly higher mAP$_{50 \rightarrow 95}$ than the “task network retraining” approach (retraining based on the representative fusion method MRFS \cite{7}) on both YOLOv5 and YOLOv11. These quantitative results demonstrate that our method maintains clear advantages on more advanced detectors.

\section{More Ablation Studies}
\label{sec:s3}

\textbf{Effectiveness of BVB}. The Basis Vector Bank (BVB) stores a set of basis vectors for generating task-specific convolutional kernel parameters. To evaluate its effectiveness, we design \textbf{Model IV}, which removes the BVB and directly predicts convolutional kernel parameters using only semantic features from downstream tasks and multimodal features. As shown in Table~\ref{tab:as_1} and Figure~\ref{labels3}, Model IV exhibits a clear performance degradation across all downstream tasks, with both quantitative and qualitative results significantly inferior to those of the full model. In contrast, the full model equipped with the BVB achieves superior and more balanced performance across multiple tasks. These results clearly demonstrate the pivotal role of the BVB in enhancing multi-task adaptability.

\textbf{Effectiveness of A2SI}. The Architecture-Adaptive Semantic Injection (A2SI) block adopts a multi-branch architecture for task-specific semantic extraction and adaptively adjusts the structure of each branch based on semantic features from different tasks. To evaluate its effectiveness, we design \textbf{Model V}, where the multi-branch structure is fixed and no longer adaptively adjusted based on task-specific semantic features. As shown in Table~\ref{tab:as_1} and Figure~\ref{labels3}, Model V exhibits a clear bias toward the object detection task and exhibits a notable decline in multi-task adaptability compared with the full model. In summary, the results confirm the essential role of the A2SI block in achieving effective multi-task adaptation.

\section{Hyperparameter Analysis}
\label{sec:s4}
The proposed method involves four key hyperparameters, namely the number of A2SI blocks $L$, the balance coefficient $\delta $ between the reward loss $\ell _{r}^n$ and the penalty loss $\ell _{p}^n$, the number of semantic extraction branches $M$ in the A2SI block, and the embedding dimension of the basis vectors $e_2$. In this work, we set $L=2$, $\delta=5$, $M=4$, and $e_2=256$. To verify the effectiveness of these hyperparameter settings on Object Detection (OD), Semantic Segmentation (Seg), and Salient Object Detection (SOD) tasks, we conduct hyperparameter analysis on the M$^3$FD \cite{10}, FMB \cite{16}, and VT5000 \cite{32} datasets.

\textbf{Impact of $\bm{L}$ on Performance}. To verify the effectiveness of the hyperparameter $L$ , we conduct a search within the range of 2 to 4. Specifically, we set three groups of hyperparameters with $L=2$, $L=3$, and $L=4$, respectively. As shown in Table \ref{tab:ha_L}, with the increase of $L$, the performance of the fused images on the semantic segmentation task gradually improves. However, the performance on the object detection task decreases to varying degrees, while that on the salient object detection task remains relatively stable. In addition, as $L$ increases, both the number of parameters and computational cost grow significantly. Therefore, considering the overall performance and efficiency, we set $L=2$ as the default value.

\textbf{Impact of $\bm {\delta} $ on Performance}. To verify the effectiveness of the hyperparameter $\delta $, we conduct a search within the range of 1 to 10. Specifically, three groups of hyperparameters are set to $\delta=1$, $\delta=5$, and $\delta=10$, respectively. As shown in Table \ref{tab:ha_delta}, when $\delta=1$, it is difficult for CLDyN to achieve balanced performance across multiple downstream tasks. Conversely, when $\delta=10$, it becomes challenging to maintain a balance between the penalty loss $\ell _{p}^n$ and the reward loss $\ell _{r}^n$, which negatively affects the optimization of all tasks. Only when $\delta=5$ does the VFN achieves relatively stable and superior performance across all downstream tasks. Therefore, we set the hyperparameter $\delta=5$ as the default value.

\begin{table}[htbp!]
	\centering
	\caption{Quantitative analysis of the hyperparameter $\delta $ across multiple downstream tasks.  The hyperparameter settings of our proposed method are highlighted in \textbf{bold}, while the best performance for each metric is marked with a {\setlength{\fboxsep}{2.3pt}\colorbox{lightred}{Red}} background.}
	\renewcommand\arraystretch{1.4}
	{\footnotesize\centerline{\tabcolsep=11.3pt
			\begin{tabular}{c|c|c|cc}
				\toprule
				\multirow{2}{*}{$\delta $} & \textbf{OD}    & \textbf{Seg}   & \multicolumn{2}{c}{\textbf{SOD}} \\
				\cline{2-5}          & mAP$_{50 \rightarrow 95} \uparrow$   & mIoU $\uparrow$  & mF$_{\beta} \uparrow$     & E$_m \uparrow$ \\
				\midrule
				1     & 0.6278  & 60.17 & 0.8123  & \cellcolor{lightred}{0.9088} \\
				\textbf{5} & \cellcolor{lightred}{0.6304} & \cellcolor{lightred}{60.34} & \cellcolor{lightred}{0.8129}  & 0.9087  \\
				10    & 0.6293  & 60.15 & 0.8125 & 0.9085  \\
				\bottomrule
	\end{tabular}}}\vspace{-3mm}
	\label{tab:ha_delta}%
\end{table}%

\begin{table}[htbp!]
	\centering
	\caption{Quantitative analysis of the hyperparameter $M$ across multiple downstream tasks.  The hyperparameter settings of our proposed method are highlighted in \textbf{bold}, while the best performance for each metric is marked with a {\setlength{\fboxsep}{2.3pt}\colorbox{lightred}{Red}} background.}
	\renewcommand\arraystretch{1.4}
	{\footnotesize\centerline{\tabcolsep=11.3pt
			\begin{tabular}{c|c|c|cc}
				\toprule
				\multirow{2}{*}{M} & \textbf{OD}    & \textbf{Seg}   & \multicolumn{2}{c}{\textbf{SOD}} \\
				\cline{2-5}          & mAP$_{50 \rightarrow 95} \uparrow$   & mIoU $\uparrow$  & mF$_{\beta} \uparrow$    & E$_m \uparrow$ \\
				\midrule
				1     & 0.6289  & 60.07 & 0.8119    & 0.9086  \\
				2     & 0.6294  & 60.15 & 0.8123   & 0.9086  \\
				\textbf{4} & \cellcolor{lightred}{0.6304} & \cellcolor{lightred}{60.34} & 0.8129   & 0.9087  \\
				6     & 0.6293  & 60.29 & \cellcolor{lightred}{0.8132} & \cellcolor{lightred}{0.9091} \\
				8     & 0.6299  & 60.23 & 0.8118  & 0.9084  \\
				\bottomrule
	\end{tabular}}}\vspace{-3mm}
	\label{tab:ha_M}%
\end{table}%

\textbf{Impact of $\bm{M}$ on Performance}. To verify the effectiveness of the hyperparameter ${M}$, we perform a search within the range of 1 to 8. Specifically, five groups of hyperparameters are set as $M=1$, $M=2$,  $M=4$, $M=6$, and $M=8$, respectively. As shown in Table \ref{tab:ha_M}, setting too few or too many semantic extraction branches weakens the adaptability of the method across multiple downstream tasks. When $M=4$, the fused images achieve the best performance on both object detection and semantic segmentation tasks, and perform comparably well on salient object detection. Therefore, we set the hyperparameter $M=4$ as the default value.

\textbf{Impact of $\bm{e_2}$ on Performance}. To verify the effectiveness of the embedding dimension ${e_2}$, we perform a search within the range of 128 to 512. Specifically, the embedding dimensions of the basis vectors are set to 128, 256, and 512, respectively. As shown in Table \ref{tab:ha_e2}, when ${e_2}=256$, the proposed method achieves the strongest adaptability across multiple downstream tasks, maintaining consistently superior performance in all tasks. Therefore, we set ${e_2}=256$ as the default configuration.

\begin{table}[htbp!]
	\centering
	\caption{Quantitative analysis of the hyperparameter ${e_2}$ across multiple downstream tasks. The hyperparameter settings of our proposed method are highlighted in \textbf{bold}, while the best performance for each metric is marked with a {\setlength{\fboxsep}{2.3pt}\colorbox{lightred}{Red}} background.}
	\renewcommand\arraystretch{1.4}
	{\footnotesize
		\setlength{\tabcolsep}{10.8pt}
		\begin{tabular}{c|c|c|cc}
			\toprule
			\multirow{2}{*}{${e_2}$} & \textbf{OD}    & \textbf{Seg} & \multicolumn{2}{c}{\textbf{SOD}} \\
			\cline{2-5}
			& mAP$_{50\rightarrow95}\uparrow$ & mIoU $\uparrow$ & mF$_{\beta}\uparrow$ & E$_m\uparrow$ \\
			\midrule
			128   & 0.6273  & 60.18 & 0.8118  & 0.9083  \\
			\textbf{256} & \cellcolor{lightred}{0.6304} & \cellcolor{lightred}{60.34} & 0.8129  & 0.9087  \\
			512   & 0.6292  & 60.25 & \cellcolor{lightred}{0.8130} & \cellcolor{lightred}{0.9092} \\
			\bottomrule
	\end{tabular}}
	\label{tab:ha_e2}
\end{table}

\begin{table}[htbp]
	\centering
	\caption{Parameter counts and computational costs of the core components in the proposed method and their corresponding ablated variants.}
	\renewcommand\arraystretch{1.4}
	{\footnotesize
		\setlength{\tabcolsep}{9.8pt}
	\begin{tabular}{c|c c|c|c}
		\toprule
		\textbf{Models} & \textbf{A2SI}  & \textbf{BVB}   & \multicolumn{1}{c|}{\textbf{Params (M)}} & \multicolumn{1}{c}{\textbf{FLOPs (G)}} \\
		\midrule
		IV    & \ding{51}    & \ding{55} & 0.11  & 1.88 \\
		V     & \ding{55}    & \ding{51}     & 0.35  & 173.04 \\
		Full  & \ding{51}    & \ding{51}    & 0.46  & 174.06 \\
		\bottomrule
	\end{tabular}}
	\label{tab:mca}%
\end{table}%

\begin{figure}[t!]
	\centering
	\includegraphics[width=0.40\textwidth]{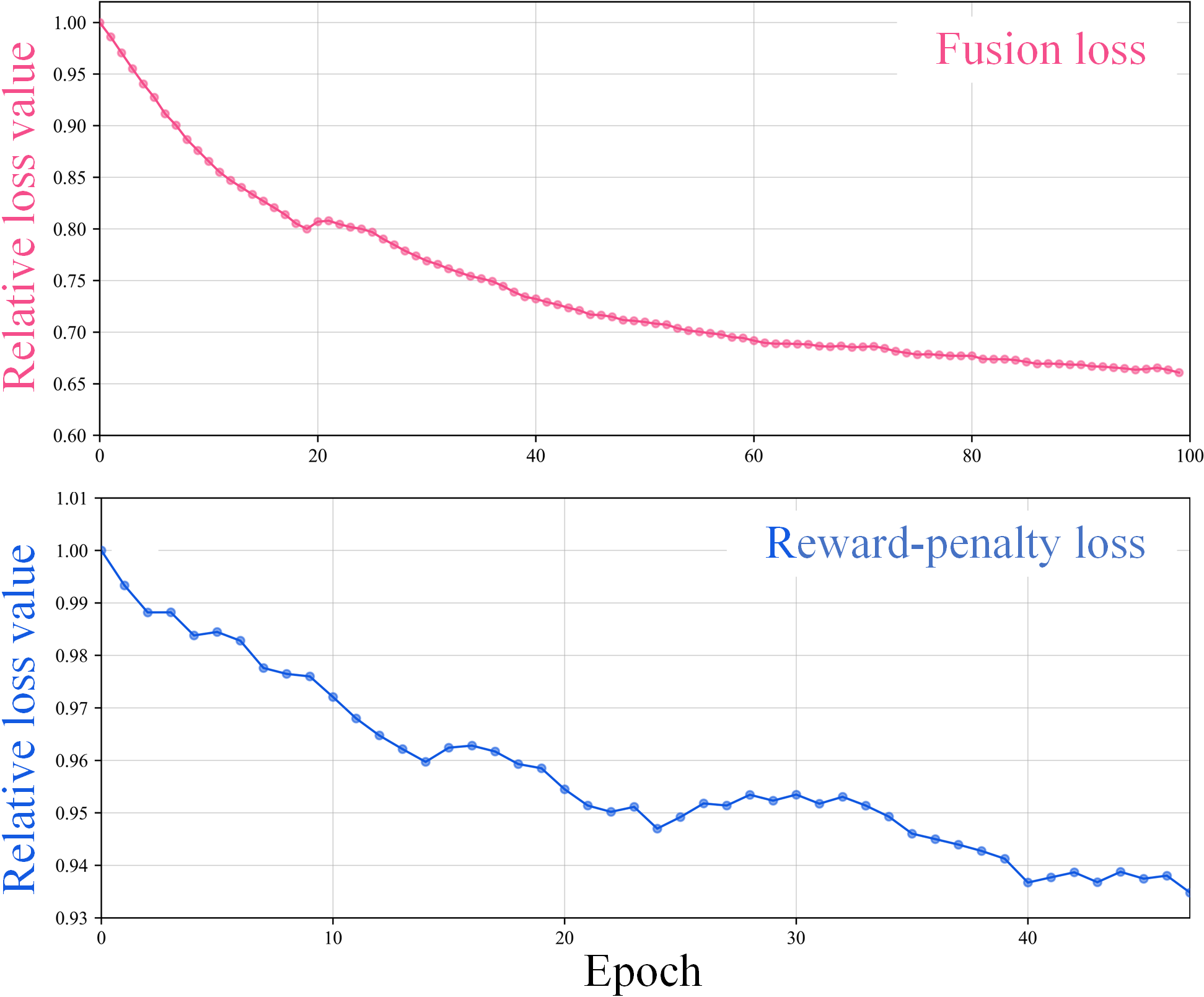}
	\caption{Training loss curves of the proposed method.}
	\label{labels4}
\end{figure}

\section{Convergence Analysis}
To verify the stable optimization and convergence of the proposed method, we visualize the fusion loss and the reward-penalty loss during training. As shown in Figure~ \ref{labels4}, both losses decrease smoothly and gradually converge without catastrophic oscillations.

\section{More Complexity Analysis}
To better analyze the complexity of the proposed method, we report the parameter counts and computational costs of each core component, as shown in Table \ref{tab:mca}.

\begin{table}[htbp]
	\centering
	\caption{Quantitative results under complex scenarios (exposure attenuation $\alpha \in [0.4, 0.8]$).}
	\renewcommand\arraystretch{1.4}
	{\footnotesize
		\setlength{\tabcolsep}{9pt}
	\begin{tabular}{c|c|c|cc}
		\toprule
		\multirow{2}{*}{\thead{\textbf{Complex} \\ \textbf{scenarios}}} & \textbf{OD}    & \textbf{Seg} & \multicolumn{2}{c}{\textbf{SOD}} \\
		\cline{2-5} & mAP$_{50\rightarrow95}\uparrow$ & mIoU $\uparrow$ & mF$_{\beta}\uparrow$ & E$_m\uparrow$ \\
		\midrule
		\ding{51}   & 0.6284  & 59.73 & 0.8100  & 0.9074  \\
		\ding{55} & 0.6304 & 60.34 & 0.8129  & 0.9087  \\
		\bottomrule
	\end{tabular}}
	\label{tab:lfw}%
\end{table}%

\section{Limitations and Future Work}
\label{sec:s5}
The proposed method assumes that the input infrared and visible images are captured under normal weather conditions. However, in open-world applications, the algorithm often encounters complex and dynamic environments such as heavy rain, low-light conditions, and severe noise, where its performance may degrade, as shown in Table \ref{tab:lfw}. To address this challenge, future research will focus on developing a multi-task-aware infrared and visible image fusion framework that remains robust under extreme weather conditions, enabling reliable performance across diverse downstream tasks. In addition to external weather factors, the aging of internal sensor components can also deteriorate image quality, leading to increased noise, pixel defects, and intensified artifacts. Future work will further explore multi-task-aware infrared and visible image fusion under complex compound degradations, aiming to enhance the robustness and generalization of the algorithm in open-world scenarios.

\end{document}